%% file: main.tex
\documentclass[10pt,twocolumn,letterpaper]{article}

\usepackage[pagenumbers]{cvpr}

\usepackage{tabularx}
\usepackage{booktabs}
\usepackage[labelsep=period,font=small]{caption}
\usepackage{tikz}
\usepackage{tikzfill}
\usepackage{pgfplots}
\usepackage{pgfplotstable}
\usepackage{standalone}
\usepackage{units}
\usepackage{multirow,makecell}

\graphicspath{{./figures/}}

\newlength{\itemwidth}

\newcolumntype{Y}{>{\centering\arraybackslash}X}
\newcolumntype{P}[1]{>{\centering\arraybackslash}p{#1}}

\def\smallcheck{\tikz\fill[scale=0.25](0,.35) -- (.25,0) -- (1,.7) -- (.25,.15) -- cycle;}
\def\crossmark{\tikz{\draw[line width=1pt](0,0.2) -- (0.2,0);\draw[line width=1pt](0,0) -- (0.2,0.2);}}
\newcommand*{\inparagraph}[1]{\medskip\noindent\textbf{#1}\hspace{0.5em}}

\usetikzlibrary{spy}
\usetikzlibrary{arrows.meta}
\usepgfplotslibrary{polar}
\pgfplotsset{compat=newest}

\definecolor{3787CF}{RGB}{55,135,207}
\definecolor{F1C232}{RGB}{241,194,50}
\definecolor{9D4761}{RGB}{157,71,97}
\definecolor{619D47}{RGB}{97,157,71}

\pgfplotscreateplotcyclelist{custompalette}{
    {3787CF!90!black,line width=1.5pt},
    {F1C232!90!black,line width=1.5pt},
    {9D4761!90!black,line width=1.5pt},
    {619D47!90!black,line width=1.5pt}
}

\pgfplotscreateplotcyclelist{custompalettethin}{
    {3787CF!90!black,line width=1.0pt},
    {F1C232!90!black,line width=1.0pt},
    {9D4761!90!black,line width=1.0pt},
    {619D47!90!black,line width=1.0pt}
}

\usepackage{listings}

\NewDocumentCommand{\headerot}{O{45} O{1em} m}{\makebox[#2][l]{\rotatebox{#1}{#3}}}%

\usepackage{stmaryrd}
\usepackage{trimclip}

\makeatletter
\DeclareRobustCommand{\shortto}{%
  \mathrel{\mathpalette\short@to\relax}%
}

\DeclareRobustCommand{\veryshortto}{%
  \mathrel{\mathpalette\veryshort@to\relax}%
}

\newcommand{\short@to}[2]{%
  \mkern2mu
  \clipbox{{.3\width} 0 0 0}{$\m@th#1\vphantom{+}{\shortrightarrow}$}%
  }

\newcommand{\veryshort@to}[2]{%
  \mkern2mu
  \clipbox{{.2\width} 0 0 0}{$\m@th#1\vphantom{+}{\shortrightarrow}$}%
  }
\makeatother

\newcommand{\usolid}[1]{%
    \tikz[remember picture, baseline=(tosolid.base)]{
        \node[inner sep=0pt, outer sep=0pt] (tosolid) {#1};
    }%
    \tikz[remember picture, overlay]{
        \draw[] ([yshift=-1.5pt]tosolid.south west) -- ([yshift=-1.5pt]tosolid.south east);
    }%
}%

\newcommand{\udotted}[1]{%
    \tikz[remember picture, baseline=(todotted.base)]{
        \node[inner sep=0pt, outer sep=0pt] (todotted) {#1};
    }%
    \tikz[remember picture, overlay]{
        \draw[densely dotted, line width=0.5] ([yshift=-1.5pt]todotted.south west) -- ([yshift=-1.5pt]todotted.south east);
    }%
}%

\newcommand{\uarrow}[1]{%
    \tikz[remember picture, baseline=(toarrow.base)]{
        \node[inner sep=0pt, outer sep=0pt] (toarrow) {#1};
    }%
    \tikz[remember picture, overlay]{
        \draw[-latex,line width=0.01cm] ([yshift=-1.5pt]toarrow.south west) -- ([yshift=-1.5pt]toarrow.south east);
    }%
}%

\definecolor{cvprblue}{rgb}{0.21,0.49,0.74}
\usepackage[pagebackref,breaklinks,colorlinks,citecolor=cvprblue]{hyperref}

\title{Benchmarking Video Frame Interpolation}

\author{
Simon Kiefhaber\\{\small TU Darmstadt, hessian.AI}\\
\and
Simon Niklaus\\{\small Adobe Research}\\
\and
Feng Liu\\{\small Adobe Research}\\
\and
Simone Schaub-Meyer\\{\small TU Darmstadt, hessian.AI}\\
}

\begin{document}

\maketitle

\begin{abstract}

Video frame interpolation, the task of synthesizing new frames in between two or more given ones, is becoming an increasingly popular research target. However, the current evaluation of frame interpolation techniques is not ideal. Due to the plethora of test datasets available and inconsistent computation of error metrics, a coherent and fair comparison across papers is very challenging. Furthermore, new test sets have been proposed as part of method papers so they are unable to provide the in-depth evaluation of a dedicated benchmarking paper. Another severe downside is that these test sets violate the assumption of linearity when given two input frames, making it impossible to solve without an oracle. We hence strongly believe that the community would greatly benefit from a benchmarking paper, which is what we propose. Specifically, we present a benchmark which establishes consistent error metrics by utilizing a submission website that computes them, provides insights by analyzing the interpolation quality with respect to various per-pixel attributes such as the motion magnitude, contains a carefully designed test set adhering to the assumption of linearity by utilizing synthetic data, and evaluates the computational efficiency in a coherent manner.

\end{abstract}

\section{Introduction}
\label{sec:intro}

\begin{figure}\centering
    \setlength{\tabcolsep}{0.0cm}
    \renewcommand{\arraystretch}{1.2}
    \newcommand{\quantTit}[1]{\makebox[0.0cm][l]{\hspace{-0.13cm}\rotatebox{75}{\scriptsize #1}}}
    \newcommand{\quantVal}[1]{\scalebox{0.83}[1.0]{$ #1 $}}
    \footnotesize
    \begin{tabularx}{\columnwidth}{@{\hspace{0.1cm}} X c @{\hspace{0.18cm}} c @{\hspace{0.18cm}} c @{\hspace{0.18cm}} c @{\hspace{0.18cm}} c @{\hspace{0.18cm}} c @{\hspace{0.18cm}} c @{\hspace{0.18cm}} c @{\hspace{0.18cm}} c @{\hspace{0.1cm}}}
        \toprule
             & \quantTit{resolution} & \quantTit{number of samples} & \quantTit{multi-frame} & \quantTit{paper occurrences} & \quantTit{submission page} & \quantTit{per-pixel metrics} & \quantTit{linear motion} & \quantTit{compute evaluation} & \quantTit{data source}
        \\ \midrule
UCF101~\cite{Soomro:2012:DHA,Liu:2017:VFS} & 256$\times$256 & 379 & \crossmark & 28 & \crossmark & \crossmark & \crossmark & \crossmark & videos
\\
Vimeo-90k~\cite{Xue:2019:VET} & 448$\times$256 & 3782 & \crossmark & 26 & \crossmark & \crossmark & \crossmark & \crossmark & videos
\\
Middlebury~\cite{Baker:2011:DBE} & 480p & 8 & \crossmark & 16 & \smallcheck & \crossmark & \crossmark & \crossmark & mixed
\\
SNU~\cite{Choi:2020:CAA} & 720p & 1240 & \crossmark & 12 & \crossmark & \crossmark & \crossmark & \crossmark & videos
\\
Xiph~\cite{Niklaus:2020:SSV} & 1K - 4K & 392 & \crossmark & 11 & \crossmark & \crossmark & \crossmark & \crossmark & videos
\\
X-TEST~\cite{Sim:2021:XVF} & 1K - 4K & 15 & \smallcheck & 8 & \crossmark & \crossmark & \crossmark & \crossmark & videos
\\
Davis~\cite{Perazzi:2016:BDE} & $-$ & $-$ & \crossmark & 5 & \crossmark & \crossmark & \crossmark & \crossmark & videos
\\
Adobe240~\cite{Su:2017:DVD} & 720p & $-$ & \smallcheck & 5 & \crossmark & \crossmark & \crossmark & \crossmark & videos
\\
HD~\cite{Bao:2019:DAV} & 1K - 2K & $-$ & \smallcheck & 3 & \crossmark & \crossmark & \crossmark & \crossmark & mixed
\\
Sintel~\cite{Butler:2012:NOS,Janai:2017:EHS} & 1024$\times$436 & 13 & \smallcheck & 3 & \crossmark & \crossmark & \crossmark & \crossmark & synthetic
\\
Ours & 1K - 4K & 666 & \smallcheck & N/A & \smallcheck & \smallcheck & \smallcheck & \smallcheck & synthetic
        \\ \bottomrule
    \end{tabularx}\vspace{-0.2cm}
    \captionof{table}{Overview of typical test datasets for frame interpolation. We list the triplet version of Vimeo-90k, occurrences are counted across 37 papers since 2018 at major computer vision venues, and missing entries denote inconsistencies between papers.}\vspace{-0.2cm}
    \label{tbl:datasets}
\end{figure}

Video frame interpolation, the synthesis of new imagery in between two or more given video frames, is becoming an increasingly popular research target. After all, frame interpolation is finding ever more creative applications such as in video compression~\cite{Wu:2018:VCT,Djelouah:2019:NIF}, video editing~\cite{Meyer:2018:DVC}, motion blur synthesis~\cite{Brooks:2019:LSM}, and many others. And not only are there more and more approaches being proposed to address frame interpolation, there is also an increasing number of datasets involved when evaluating them. There are earlier datasets like the interpolation category of the Middlebury benchmark for optical flow estimation~\cite{Baker:2011:DBE}, as well as more recent datasets like X-TEST~\cite{Sim:2021:XVF} which were introduced to analyze new settings such as higher resolution footage or multiple in-between frames. Please see \Cref{tbl:datasets} for an overview.

The diversity in test datasets leads to a division in the frame evaluation space since it is unreasonable to expect new papers to analyze their proposed approach on all of them. Additionally, error metrics often do not match across papers even though they may have used the same test set due to different compute paradigms (for example, using floating point versus quantized integer representation), different ways to compute error metrics (computing the structural similarity with or without correlating the color channels), or the same name sometimes referring to different dataset variants (Vimeo-90k triplet versus Vimeo-90k septuplet). As a result, it is increasingly difficult to compare approaches.

\begin{figure*}\centering
    \setlength{\tabcolsep}{0.05cm}
    \setlength{\itemwidth}{4.27cm}
    \hspace*{-\tabcolsep}\begin{tabular}{cc}
            \begin{tikzpicture}[spy using outlines={3787CF, magnification=4, width={\itemwidth - 0.06cm}, height=3.13cm, connect spies,
    every spy in node/.append style={line width=0.06cm}}]
                \node [inner sep=0.0cm] {\includegraphics[width=\itemwidth]{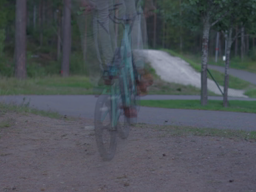}};
                \spy [every spy on node/.append style={line width=0.06cm}, spy connection path={\draw[line width=0.06cm] (tikzspyonnode) -- (tikzspyinnode);}] on (-0.1,-0.75) in node at (4.39,0.0);
            \end{tikzpicture}
        &
            \begin{tikzpicture}[spy using outlines={3787CF, magnification=4, width={\itemwidth - 0.06cm}, height=3.13cm, connect spies,
    every spy in node/.append style={line width=0.06cm}}]
                \node [inner sep=0.0cm] {\includegraphics[width=\itemwidth]{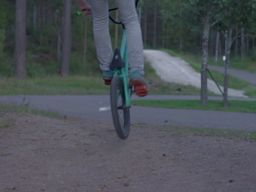}};
                \draw [draw=F1C232, line width=0.01cm] (-0.05+0.05,-0.66-0.04) -- (-0.05-0.05,-0.66+0.04);
                \draw [draw=F1C232, line width=0.01cm,-{Triangle[length=0.05cm, width=0.05cm]}] (-0.05,-0.66) -- (0.075,-0.49) node [midway, sloped, above, fill=white, inner sep=0.07cm, shift={(-0.02cm,0.02cm)}, scale=0.15] {$F_{1 \shortto 2}$};
                \draw [draw=F1C232, line width=0.01cm,-{Triangle[length=0.05cm, width=0.05cm]}] (-0.05,-0.66) -- (-0.32,-1.02) node [midway, sloped, above, fill=white, inner sep=0.07cm, shift={(0.14cm,0.02cm)}, scale=0.15] {$F_{1 \shortto 0}$};
                \spy [every spy on node/.append style={line width=0.06cm}, spy connection path={\draw[line width=0.06cm] (tikzspyonnode) -- (tikzspyinnode);}] on (-0.1,-0.75) in node at (4.39,0.0);
            \end{tikzpicture}
        \\
            \footnotesize overlaid frame triplet of a bicycle in motion
        &
            \footnotesize center frame with the annotated optical flow
        \\
    \end{tabular}\vspace{-0.2cm}
    \caption{Existing test datasets for frame interpolation violate the constraint of linearity. Like in this example from Vimeo-90k of a jumping bicycle, it is hence impossible at times to infer the alleged ground truth from the inputs since it is not temporally centered.}\vspace{-0.3cm}
    \label{fig:nonlinear}
\end{figure*}

Furthermore, existing test datasets often do not provide deep insights since they are only part of a paper that focuses on proposing a new interpolation method. As such, these test datasets do not enjoy the benefit of an in-depth analysis that a dedicated benchmarking paper could provide. For example, the larger the inter-frame motion the more difficult it is to interpolate but there is no existing thorough evaluation of the error with respect to the underlying motion magnitude. Similarly, certain areas are more difficult to interpolate since they are only visible in one, or even none, of the input frames but no paper has analyzed this aspect.

Another severe downside of existing test sets, as shown in \Cref{fig:nonlinear}, is that they violate the constraint of linearity when given two input frames since they sample adjacent frames from off-the-shelf videos. However, given two frames one can only reasonably expect the motion to be linear. Any other assumption would require an oracle when synthesizing the in-between frames. As such, if the ground truth does not follow this assumption then it fails at being a ground truth. Looking at it from an optimization angle, when given two points and being asked to fit a model then any model other than a linear one would be underdetermined and hence yield an infinite number of solutions. We are thus not intentionally limiting the motion in our benchmark, we are just implementing the only valid solution.

Although independent of a new dataset, the evaluation of the computational performance of frame interpolation is far from ideal. Sometimes the evaluation of the timings is imprecise due to ignoring the asynchronous nature of modern graphics processors, thus accidentally measuring how long it takes to issue the compute kernels and not the actual execution time. Also, sometimes the evaluation is copied from another paper and only the timings for the proposed approach are being measured, which does not allow for meaningful comparisons since differences in hardware and even software versions can have a significant impact.

Given these circumstances, we strongly believe that the frame interpolation community would greatly benefit from a benchmarking paper. Specifically, we propose a benchmark which establishes consistent error metrics by utilizing a submission website that computes them, provides deep insights by analyzing the interpolation quality with respect to various attributes, adheres to the constraint of linear motion by utilizing synthetic data, and evaluates the computational efficiency in a unified and hence coherent manner.

Our benchmark consists of 666 nonuplets at up to 4K resolution, which are synthetically rendered and thus allow us to conform to the constraint of linear motion. The synthetic data further enables us to extract auxiliary information such as the ground truth motion magnitude, giving us the opportunity to evaluate the interpolation quality with respect to various per-pixel attributes. Users of the benchmark are only given the first and last frame in each sequence though, together with a pip-installable library that uploads the results to an evaluation server. This is a common practice in benchmarks~\cite{Cordts:2016:CDS,Geiger:2012:AWR,Butler:2012:NOS,Kondermann:2016:HBS,Baker:2011:DBE,Everingham:2010:PVO} to ensure consistent results and to prevent accidental supervision on the test data. Our benchmark already includes 21 methods, and as a byproduct, we will provide a convenient way for others to run these methods to conveniently facilitate comparisons.

In doing so, we will answer some key questions such as how significant the impact of non-linear motion in existing test datasets is, whether there is a domain gap between our synthetic data and existing test datasets, and how to better evaluate multi-frame consistency. We also do not shy away from being explicit about the limitations of our benchmark. For example, we focus on frame interpolation methods that take two frames as input, thus omitting areas such as quadratic frame interpolation~\cite{Xu:2019:QVI} or frame interpolation of event footage~\cite{Tulyakov:2021:TLE,Tulyakov:2022:TLE}. Furthermore, since we analyze the interpolation quality with respect to various per-pixel attributes such as the motion magnitude, we are unfortunately unable to incorporate more advanced error metrics due to their inability to handle sparse inputs~\cite{Wang:2004:IQA,Zhang:2018:UED,Hou:2022:PQM}.

\section{Related Work}
\label{sec:related}

By using skipped frames as ground truth, any video can in theory be used to evaluate frame interpolation methods. As a result and as summarized in \Cref{tbl:datasets}, there is a wide variety of datasets to choose from. The most commonly used ones are Vimeo-90k~\cite{Xue:2019:VET}, UCF101~\cite{Soomro:2012:DHA,Liu:2017:VFS}, and the interpolation set of the Middlebury benchmark~\cite{Baker:2011:DBE} even though they are limited by their low resolution and by only providing a single ground truth (note that the Middlebury benchmark has stopped accepting submissions). In contrast, newer datasets operate at a higher resolution like Xiph~\cite{Niklaus:2020:SSV}, provide different categories based on difficulty like SNU~\cite{Choi:2020:CAA}, and evaluate multi-frame interpolation like X-TEST~\cite{Sim:2021:XVF}.

However, the existing test sets do not have a submission page which leads to inconsistencies across papers due to different paradigms when computing the error metrics (with or without quantization prior to computing the metrics, or different ways of computing the SSIM~\cite{Wang:2004:IQA}). Furthermore, the lack of a leaderboard makes it difficult to keep up with the state of the art since one has to go through the most-recent arXiv papers instead (and concurrent papers do not compare to each other, amplifying this problem). Other limitations of these test sets include the lack of an in-depth evaluation, the violation of the constraint of linearity, and the lack of an evaluation of the computational efficiency. In comparison, our benchmark addresses these limitations.

\begin{figure}\centering
    \setlength{\tabcolsep}{0.0cm}
    \renewcommand{\arraystretch}{1.2}
    \newcommand{\quantTit}[1]{\multicolumn{2}{c}{\scriptsize #1}}
    \newcommand{\quantSec}[1]{\scriptsize #1}
    \newcommand{\quantInd}[1]{\scriptsize #1}
    \newcommand{\quantVal}[1]{\scalebox{0.83}[1.0]{$ #1 $}}
    \newcommand{\quantFirst}[1]{\usolid{\scalebox{0.83}[1.0]{$ #1 $}}}
    \newcommand{\quantSecond}[1]{\udotted{\scalebox{0.83}[1.0]{$ #1 $}}}
    \footnotesize
    \begin{tabularx}{\columnwidth}{@{\hspace{0.1cm}} X P{1.55cm} @{\hspace{-0.1cm}} P{1.55cm} P{1.55cm} @{\hspace{-0.1cm}} P{1.55cm}}
        \toprule
            & \quantTit{Vimeo-90k} & \quantTit{Xiph - 1K}
        \\ \cmidrule(l{2pt}r{2pt}){2-3} \cmidrule(l{2pt}r{2pt}){4-5}
            & {\vspace{-0.29cm} \scriptsize all \linebreak pixels} & {\vspace{-0.29cm} \scriptsize 97\% \linebreak most-linear} & {\vspace{-0.29cm} \scriptsize all \linebreak pixels} & {\vspace{-0.29cm} \scriptsize 97\% \linebreak most-linear}
        \\ \midrule
ABME~\cite{Park:2021:ABM} & \quantVal{33.83}\makebox[0cm]{\raisebox{0.065cm}{\hspace{0.92cm}\scalebox{0.6}[0.9]{\tiny\uarrow{\thinspace\thinspace\thinspace\thinspace\thinspace + $1.27$ \thinspace\thinspace\thinspace\thinspace\thinspace\thinspace}}}} & \quantVal{35.10} & \quantVal{34.90}\makebox[0cm]{\raisebox{0.065cm}{\hspace{0.92cm}\scalebox{0.6}[0.9]{\tiny\uarrow{\thinspace\thinspace\thinspace\thinspace\thinspace + $0.95$ \thinspace\thinspace\thinspace\thinspace\thinspace\thinspace}}}} & \quantVal{35.85}
\\
AMT-G~\cite{Li:2023:AMT} & \quantVal{34.15}\makebox[0cm]{\raisebox{0.065cm}{\hspace{0.92cm}\scalebox{0.6}[0.9]{\tiny\uarrow{\thinspace\thinspace\thinspace\thinspace\thinspace + $1.19$ \thinspace\thinspace\thinspace\thinspace\thinspace\thinspace}}}} & \quantVal{35.34} & \quantVal{35.90}\makebox[0cm]{\raisebox{0.065cm}{\hspace{0.92cm}\scalebox{0.6}[0.9]{\tiny\uarrow{\thinspace\thinspace\thinspace\thinspace\thinspace + $1.05$ \thinspace\thinspace\thinspace\thinspace\thinspace\thinspace}}}} & \quantVal{36.95}
\\
CtxSyn~\cite{Niklaus:2018:CAS} & \quantVal{32.42}\makebox[0cm]{\raisebox{0.065cm}{\hspace{0.92cm}\scalebox{0.6}[0.9]{\tiny\uarrow{\thinspace\thinspace\thinspace\thinspace\thinspace + $1.29$ \thinspace\thinspace\thinspace\thinspace\thinspace\thinspace}}}} & \quantVal{33.71} & \quantVal{34.31}\makebox[0cm]{\raisebox{0.065cm}{\hspace{0.92cm}\scalebox{0.6}[0.9]{\tiny\uarrow{\thinspace\thinspace\thinspace\thinspace\thinspace + $1.06$ \thinspace\thinspace\thinspace\thinspace\thinspace\thinspace}}}} & \quantVal{35.37}
\\
DAIN~\cite{Bao:2019:DAV} & \quantVal{32.49}\makebox[0cm]{\raisebox{0.065cm}{\hspace{0.92cm}\scalebox{0.6}[0.9]{\tiny\uarrow{\thinspace\thinspace\thinspace\thinspace\thinspace + $1.31$ \thinspace\thinspace\thinspace\thinspace\thinspace\thinspace}}}} & \quantVal{33.80} & \quantVal{34.74}\makebox[0cm]{\raisebox{0.065cm}{\hspace{0.92cm}\scalebox{0.6}[0.9]{\tiny\uarrow{\thinspace\thinspace\thinspace\thinspace\thinspace + $1.19$ \thinspace\thinspace\thinspace\thinspace\thinspace\thinspace}}}} & \quantVal{35.93}
\\
FLDR~\cite{Nottebaum:2022:EFE} & \quantVal{31.02}\makebox[0cm]{\raisebox{0.065cm}{\hspace{0.92cm}\scalebox{0.6}[0.9]{\tiny\uarrow{\thinspace\thinspace\thinspace\thinspace\thinspace + $1.26$ \thinspace\thinspace\thinspace\thinspace\thinspace\thinspace}}}} & \quantVal{32.28} & \quantVal{33.07}\makebox[0cm]{\raisebox{0.065cm}{\hspace{0.92cm}\scalebox{0.6}[0.9]{\tiny\uarrow{\thinspace\thinspace\thinspace\thinspace\thinspace + $1.14$ \thinspace\thinspace\thinspace\thinspace\thinspace\thinspace}}}} & \quantVal{34.21}
\\
M2M~\cite{Hu:2022:MMS} & \quantVal{33.32}\makebox[0cm]{\raisebox{0.065cm}{\hspace{0.92cm}\scalebox{0.6}[0.9]{\tiny\uarrow{\thinspace\thinspace\thinspace\thinspace\thinspace + $1.21$ \thinspace\thinspace\thinspace\thinspace\thinspace\thinspace}}}} & \quantVal{34.53} & \quantVal{35.14}\makebox[0cm]{\raisebox{0.065cm}{\hspace{0.92cm}\scalebox{0.6}[0.9]{\tiny\uarrow{\thinspace\thinspace\thinspace\thinspace\thinspace + $1.11$ \thinspace\thinspace\thinspace\thinspace\thinspace\thinspace}}}} & \quantVal{36.25}
\\
SoftSplat~\cite{Niklaus:2020:SSV} & \quantVal{33.76}\makebox[0cm]{\raisebox{0.065cm}{\hspace{0.92cm}\scalebox{0.6}[0.9]{\tiny\uarrow{\thinspace\thinspace\thinspace\thinspace\thinspace + $1.25$ \thinspace\thinspace\thinspace\thinspace\thinspace\thinspace}}}} & \quantVal{35.01} & \quantVal{35.72}\makebox[0cm]{\raisebox{0.065cm}{\hspace{0.92cm}\scalebox{0.6}[0.9]{\tiny\uarrow{\thinspace\thinspace\thinspace\thinspace\thinspace + $1.17$ \thinspace\thinspace\thinspace\thinspace\thinspace\thinspace}}}} & \quantVal{36.89}
\\
SplatSyn~\cite{Niklaus:2023:SSV} & \quantVal{32.86}\makebox[0cm]{\raisebox{0.065cm}{\hspace{0.92cm}\scalebox{0.6}[0.9]{\tiny\uarrow{\thinspace\thinspace\thinspace\thinspace\thinspace + $1.29$ \thinspace\thinspace\thinspace\thinspace\thinspace\thinspace}}}} & \quantVal{34.15} & \quantVal{34.41}\makebox[0cm]{\raisebox{0.065cm}{\hspace{0.92cm}\scalebox{0.6}[0.9]{\tiny\uarrow{\thinspace\thinspace\thinspace\thinspace\thinspace + $0.96$ \thinspace\thinspace\thinspace\thinspace\thinspace\thinspace}}}} & \quantVal{35.37}
\\
UPR-Net-L~\cite{Jin:2023:UPR} & \quantVal{34.08}\makebox[0cm]{\raisebox{0.065cm}{\hspace{0.92cm}\scalebox{0.6}[0.9]{\tiny\uarrow{\thinspace\thinspace\thinspace\thinspace\thinspace + $1.23$ \thinspace\thinspace\thinspace\thinspace\thinspace\thinspace}}}} & \quantVal{35.31} & \quantVal{35.97}\makebox[0cm]{\raisebox{0.065cm}{\hspace{0.92cm}\scalebox{0.6}[0.9]{\tiny\uarrow{\thinspace\thinspace\thinspace\thinspace\thinspace + $1.03$ \thinspace\thinspace\thinspace\thinspace\thinspace\thinspace}}}} & \quantVal{37.00}
\\
XVFI~\cite{Sim:2021:XVF} & \quantVal{30.54}\makebox[0cm]{\raisebox{0.065cm}{\hspace{0.92cm}\scalebox{0.6}[0.9]{\tiny\uarrow{\thinspace\thinspace\thinspace\thinspace\thinspace + $1.18$ \thinspace\thinspace\thinspace\thinspace\thinspace\thinspace}}}} & \quantVal{31.72} & \quantVal{32.97}\makebox[0cm]{\raisebox{0.065cm}{\hspace{0.92cm}\scalebox{0.6}[0.9]{\tiny\uarrow{\thinspace\thinspace\thinspace\thinspace\thinspace + $1.09$ \thinspace\thinspace\thinspace\thinspace\thinspace\thinspace}}}} & \quantVal{34.06}
        \\ \bottomrule
    \end{tabularx}\vspace{-0.2cm}
    \captionof{table}{Ignoring just three percent of the most non-linear pixels of common test datasets significantly alters the $\text{PSNR}^{\ast}$. This begs the question of how insightful such tests on non-linear data are.}\vspace{-0.3cm}
    \label{tbl:linearity}
\end{figure}

While the typical frame interpolation configuration receives two frames as input, there is also insightful research outside of this common setup. This includes areas such as quadratic frame interpolation~\cite{Xu:2019:QVI,Liu:2020:EQV} and frame interpolation on event footage~\cite{Tulyakov:2021:TLE,Tulyakov:2022:TLE}. However, addressing of these domains with a single benchmark would be far too ambitious, we are hence unable to explore these realms.

Typical error metrics include pixel-wise expressions such as the peak signal to noise ratio~(PSNR) and the interpolation error~(IE)~\cite{Baker:2011:DBE}. However, while taking the average of per-frame PSNRs is most-frequently used in the frame interpolation literature, it is not without flaws~\cite{Keles:2021:CPS}. We thus instead compute the PSNR of a dataset by taking the logarithm of averaged mean square errors. To avoid confusions we denote this definition as $\text{PSNR}^{\ast}$ throughout this paper and we extend our discussion of this topic in \Cref{sec:metrics}.

In the realm of patch-wise metrics, popular choices include the structural similarity~(SSIM)~\cite{Wang:2004:IQA}, the learned perceptual image patch similarity~(LPIPS)~\cite{Zhang:2018:UED} and its video variants~\cite{Danier:2022:FBV}, the frame interpolation perceptual similarity~(VFIPS)~\cite{Hou:2022:PQM}, as well as the perceptual frame interpolation quality metric~(PFIQM)~\cite{Yang:2008:NOQ}. While these error metrics can provide rich insights, we refrain from using them in our benchmark since it would prevent us from computing error metrics with respect to various per-pixel attributes such as the motion magnitude. More details on this in \Cref{sec:metrics}.

\begin{figure}\centering
    \setlength{\tabcolsep}{0.0cm}
    \renewcommand{\arraystretch}{1.2}
    \newcommand{\quantTit}[1]{\multicolumn{2}{c}{\scriptsize #1}}
    \newcommand{\quantSec}[1]{\scriptsize #1}
    \newcommand{\quantInd}[1]{\scriptsize #1}
    \newcommand{\quantVal}[1]{\scalebox{0.83}[1.0]{$ #1 $}}
    \newcommand{\quantFirst}[1]{\usolid{\scalebox{0.83}[1.0]{$ #1 $}}}
    \newcommand{\quantSecond}[1]{\udotted{\scalebox{0.83}[1.0]{$ #1 $}}}
    \footnotesize
    \begin{tabularx}{\columnwidth}{@{\hspace{0.1cm}} X  P{1.12cm} @{\hspace{-0.31cm}} P{1.85cm} P{1.12cm} @{\hspace{-0.31cm}} P{1.85cm}}
        \toprule
            & \quantTit{Middlebury} & \quantTit{Vimeo-90k}
        \\ \cmidrule(l{2pt}r{2pt}){2-3} \cmidrule(l{2pt}r{2pt}){4-5}
            & \quantSec{$\text{PSNR}^{\ast}$} \linebreak \quantInd{$\uparrow$} & {\vspace{-0.29cm} \scriptsize absolute \linebreak change} & \quantSec{$\text{PSNR}^{\ast}$} \linebreak \quantInd{$\uparrow$} & {\vspace{-0.29cm} \scriptsize absolute \linebreak change}
        \\ \midrule
retrained SepConv++ & \quantVal{34.65} & \quantVal{-} & \quantVal{32.18} & \quantVal{-}
\\
with 99\% most linear & \quantFirst{35.26} & \quantVal{\text{+ } 0.61 \text{ dB}} & \quantFirst{32.26} & \quantVal{\text{+ } 0.08 \text{ dB}}
\\
with 98\% most linear & \quantVal{35.16} & \quantVal{\text{+ } 0.51 \text{ dB}} & \quantVal{32.16} & \quantVal{\text{\scalebox{1.6}[1.0]{-} } 0.02 \text{ dB}}
        \\ \bottomrule
    \end{tabularx}\vspace{-0.2cm}
    \captionof{table}{Non-linear data not only affects the evaluation but also the training, ignoring the most non-linear pixels can improve the resulting model. We leave the exploration of this to future work.}\vspace{-0.3cm}
    \label{tbl:lintrain}
\end{figure}

Coming back to the violation of the constraint of linear motion, we have found that ignoring just three percent of the most non-linear pixels of common test datasets significantly alters the $\text{PSNR}^{\ast}$ as shown in \Cref{tbl:linearity}. To do so, we used GMA~\cite{Jiang:2021:LEH} to estimate the motion from the ground truth to the two input frames and took the flow deviation as a proxy for non-linearity. Our findings beg the question of how insightful evaluating on non-linear data actually is. Furthermore, non-linear data not only affects the evaluation but also the training. As shown in \Cref{tbl:lintrain}, ignoring the most non-linear pixels can improve the resulting model. We leave the exploration of this phenomenon to future work.

For more information on the topic of frame interpolation, we kindly refer to existing in-depth surveys~\cite{Dong:2023:VFI}.

\section{Dataset Generation}
\label{sec:dataset}

\begin{figure*}
    \includegraphics[]{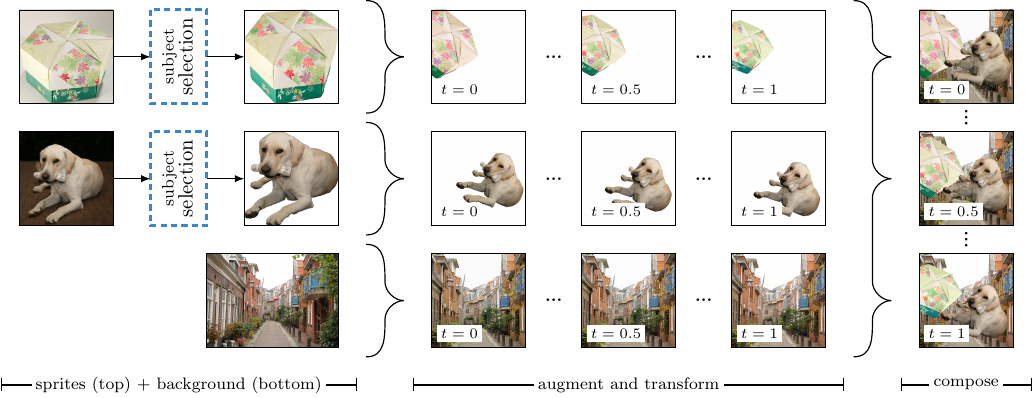}\vspace{-0.2cm}
    \caption{An overview of our dataset generation pipeline where we compose a curated set of sprites from OpenImages~\cite{Kuznetsova:2020:OID} on top of pictures from our personal collection, subject to motion from random homography transforms that strictly follow the constraint of linearity. To model photometric inconsistencies, each image is randomly augmented (notice the hue shift of the box in the first row).}\vspace{-0.2cm}
    \label{fig:datageneration}
\end{figure*}

To obtain a dataset which complies with the constraint of linear motion, we have opted for synthetic data. However, to minimize the domain gap between real and purely synthetic videos, we refrain from rendering 3D environments. We instead resort to a 2D approach that composes artificial data using real-world imagery. Specifically, we extract a set of sprites from OpenImages~\cite{Kuznetsova:2020:OID} and compose them on top of photos from our own collection as shown in \Cref{fig:datageneration}. We generated 666 sequences with two input and seven ground truth frames at a 4096$\times$2048 resolution in this way and we use anti-aliased resizing to obtain smaller resolutions.

\begin{figure}\centering
    \vspace{-0.18cm}\hspace{-0.13cm}\includegraphics[]{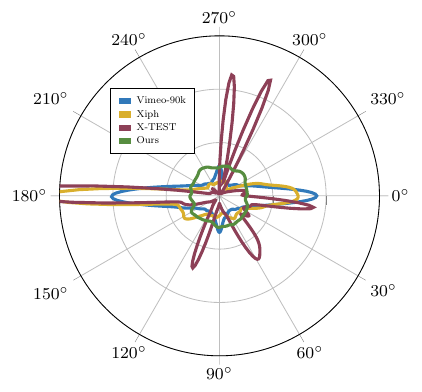}\vspace{-0.25cm}
    \caption{A histogram of the motion angle of common test data in comparison to our benchmarking dataset, demonstrating that our data is not subject to the typical bias towards horizontal motion.}\vspace{-0.2cm}
    \label{fig:histangle}
\end{figure}

To extract the sprites, we utilize OpenImages~\cite{Kuznetsova:2020:OID} where the width and height are greater than 1024 pixels. We make use of ``select subject'' from Photoshop to isolate the salient object. In case of multiple selections, we only consider the largest connected component. Moreover, the resulting sprites greatly vary in aesthetic quality which we improved through heuristic filtering rules. However, the majority of this filtering came down to laborious manual inspection. In summary, we curated 839 sprites which we randomly compose on top of 257 photos from our personal collection.

To provide the sprites as well as the background with motion, we resort to random homography transforms that strictly follow the constraint of linearity. As shown in \Cref{fig:histangle}, a welcome side-effect of this random synthetic approach to data generation is that there is no bias with respect to the direction of the motion, which is in stark contrast to existing test datasets that utilize off-the-shelf videos.

\begin{figure}\centering
    \hspace{-0.13cm}\includegraphics[]{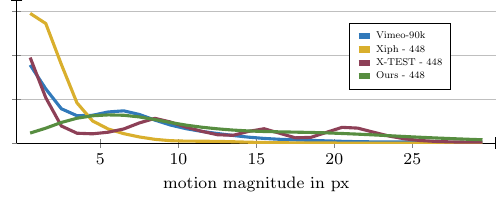}\vspace{-0.2cm}
    \caption{A histogram of the motion magnitude of common test sets in comparison to our proposed benchmarking dataset, where we resized all datasets to a width of 448 pixels to make them comparable to the size and hence magnitude of Vimeo-90k~\cite{Xue:2019:VET}.}\vspace{-0.2cm}
    \label{fig:histmag}
\end{figure}

In terms of motion magnitude, we utilized motions larger than what is typical in existing test datasets to ensure that the benchmark remains challenging for the foreseeable future. A histogram of the motion magnitude of our dataset in comparison to commonly used ones is shown in \Cref{fig:histmag}. Note that we used GMA~\cite{Jiang:2021:LEH} to estimate the motion for the other datasets which is just a rough approximation.

Lastly, we model photometric inconsistencies by augmenting the sprites as well as the background in each sequence. Specifically, we first unprocess~\cite{Brooks:2019:UIL} each image before applying various random camera pipelines through Lightroom. We do this once for the each sprite and background at $t = 0$ and at $t = 1$, and perform lerping to obtain the sprites and background for the in-between times.

Notably and as summarized in \Cref{tbl:datasets}, our dataset is synthetic while existing test sets are predominantly from in-the-wild videos. This begs the question of whether there is a domain gap between existing test sets and our synthetic one. We analyze this in \Cref{tbl:commonrank} and \Cref{tbl:commoncorr}, where we determine the performance of representative methods across various datasets before analyzing the Spearman correlation. We find that the results on Vimeo-90k, Xiph, and our benchmark correlate with each other but X-TEST does not. This also aligns with the outliers on X-TEST in terms of motion angle and magnitude depicted in \Cref{fig:histangle} and \Cref{fig:histmag}.

\section{Evaluation Metrics}
\label{sec:metrics}

The peak signal to noise ratio~(PSNR) is the most common pixel-wise error metric in the frame interpolation literature, which for a set of $N$ images is commonly defined as the average of the per-frame peak signal to noise ratios:

\begin{equation}
    \label{eq:psnr}
    \text{PSNR} = \frac{1}{N} \sum -10 \cdot \log_{10} \left( \text{MSE}_i \right)\ ,
\end{equation}

where $\text{MSE}_i$ is the mean squared error of the $i$-th image and we assume that the image intensities are in $\left[ 0, 1 \right]$. However, this is not the only way of computing the PSNR and it is arguably flawed~\cite{Keles:2021:CPS}. We hence instead use another way of computing the PSNR which is defined as:

\begin{equation}
    \label{eq:psnrx}
    \text{PSNR}^{\ast} = -10 \cdot \log_{10} \left( \frac{1}{M} \sum \text{SE}_k \right)\ ,
\end{equation}
where $M$ is the number of evaluated pixels, $\text{SE}_k$ is the squared error of the $k$-th pixel, and we are taking the logarithm after averaging $\text{SE}_k$. But why does this even matter? Consider a masked evaluation which selects one percent of one image and ninety-nine percent of another image, quite the imbalance in terms of contributions. However, the usual definition of the PSNR ignores the size of these masks since it will average the individual PSNR of each. In contrast, the $\text{PSNR}^{\ast}$ is not subject to this downside and we kindly refer to the supplementary and to~\cite{Keles:2021:CPS} for more details.

\begin{figure}\centering
    \setlength{\tabcolsep}{0.0cm}
    \renewcommand{\arraystretch}{1.2}
    \newcommand{\quantTit}[1]{\scriptsize #1}
    \newcommand{\quantSec}[1]{\scriptsize #1}
    \newcommand{\quantVal}[1]{\scalebox{0.83}[1.0]{$ #1 $}}
    \newcommand{\quantFirst}[1]{\usolid{\scalebox{0.83}[1.0]{$ #1 $}}}
    \newcommand{\quantSecond}[1]{\udotted{\scalebox{0.83}[1.0]{$ #1 $}}}
    \footnotesize
    \begin{tabularx}{\columnwidth}{@{\hspace{0.1cm}} X P{1.5cm} P{1.5cm} P{1.5cm} P{1.5cm}}
        \toprule
             & \quantTit{Vimeo-90k} & \quantTit{Xiph - 448} & \quantTit{X-TEST - 448} & \quantTit{Ours - 448}
        \\ \cmidrule(l{2pt}r{2pt}){2-2} \cmidrule(l{2pt}r{2pt}){3-3} \cmidrule(l{2pt}r{2pt}){4-4} \cmidrule(l{2pt}r{2pt}){5-5}
             & \quantSec{rank / $\text{PSNR}^{\ast}$} & \quantSec{rank / $\text{PSNR}^{\ast}$} & \quantSec{rank / $\text{PSNR}^{\ast}$} & \quantSec{rank / $\text{PSNR}^{\ast}$}
        \\ \midrule
ABME~\cite{Park:2021:ABM} & \quantVal{\hphantom{0}3} / \quantVal{33.83} & \quantVal{\hphantom{0}6} / \quantVal{34.86} & \quantVal{\hphantom{0}6} / \quantVal{29.57} & \quantVal{\hphantom{0}3} / \quantVal{30.12}
\\
AMT-G~\cite{Li:2023:AMT} & \quantVal{\hphantom{0}1} / \quantFirst{34.15} & \quantVal{\hphantom{0}3} / \quantVal{36.75} & \quantVal{10} / \quantVal{19.93} & \quantVal{\hphantom{0}2} / \quantVal{30.53}
\\
CtxSyn~\cite{Niklaus:2018:CAS} & \quantVal{\hphantom{0}8} / \quantVal{32.42} & \quantVal{\hphantom{0}7} / \quantVal{34.54} & \quantVal{\hphantom{0}2} / \quantVal{31.80} & \quantVal{\hphantom{0}5} / \quantVal{29.37}
\\
DAIN~\cite{Bao:2019:DAV} & \quantVal{\hphantom{0}7} / \quantVal{32.49} & \quantVal{\hphantom{0}5} / \quantVal{35.48} & \quantVal{\hphantom{0}7} / \quantVal{28.91} & \quantVal{\hphantom{0}8} / \quantVal{27.99}
\\
FLDR~\cite{Nottebaum:2022:EFE} & \quantVal{\hphantom{0}9} / \quantVal{31.02} & \quantVal{\hphantom{0}9} / \quantVal{32.79} & \quantVal{\hphantom{0}8} / \quantVal{28.36} & \quantVal{\hphantom{0}9} / \quantVal{23.52}
\\
M2M~\cite{Hu:2022:MMS} & \quantVal{\hphantom{0}5} / \quantVal{33.32} & \quantVal{\hphantom{0}4} / \quantVal{35.72} & \quantVal{\hphantom{0}5} / \quantVal{29.92} & \quantVal{\hphantom{0}6} / \quantVal{29.09}
\\
SoftSplat~\cite{Niklaus:2020:SSV} & \quantVal{\hphantom{0}4} / \quantVal{33.76} & \quantVal{\hphantom{0}2} / \quantVal{36.82} & \quantVal{\hphantom{0}3} / \quantVal{31.26} & \quantVal{\hphantom{0}1} / \quantFirst{30.93}
\\
SplatSyn~\cite{Niklaus:2023:SSV} & \quantVal{\hphantom{0}6} / \quantVal{32.86} & \quantVal{\hphantom{0}8} / \quantVal{34.09} & \quantVal{\hphantom{0}1} / \quantFirst{32.79} & \quantVal{\hphantom{0}7} / \quantVal{28.88}
\\
UPR-Net-L~\cite{Jin:2023:UPR} & \quantVal{\hphantom{0}2} / \quantVal{34.08} & \quantVal{\hphantom{0}1} / \quantFirst{37.11} & \quantVal{\hphantom{0}4} / \quantVal{30.28} & \quantVal{\hphantom{0}4} / \quantVal{29.68}
\\
XVFI~\cite{Sim:2021:XVF} & \quantVal{10} / \quantVal{30.54} & \quantVal{10} / \quantVal{32.03} & \quantVal{\hphantom{0}9} / \quantVal{28.22} & \quantVal{10} / \quantVal{23.41}
        \\ \bottomrule
    \end{tabularx}\vspace{-0.2cm}
    \captionof{table}{Evaluating the performance of representative interpolation methods across various datasets, where we resized all datasets to a common width of 448 pixels to make them comparable to the size and hence motion magnitude of Vimeo-90k~\cite{Xue:2019:VET}.}\vspace{-0.3cm}
    \label{tbl:commonrank}
\end{figure}

\begin{figure}
    \includegraphics[]{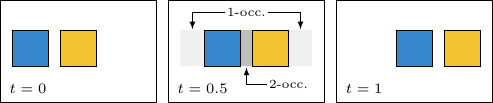}\vspace{-0.2cm}
    \caption{Occlusions are a well-known challenge in frame interpolation, and pixels can be occluded in one (1-occ.) or even in both input frames (2-occ.). Thanks to the synthetic nature of our data, we can pinpoint these areas and assess them separately.}\vspace{-0.3cm}
    \label{fig:occtypes}
\end{figure}

Thanks to the synthetic nature of our data, we know the true optical flow which allows us to analyze the $\text{PSNR}^{\ast}$ for regions of interest. Specifically, a well-known challenge in frame interpolation are occlusions and we can use the consistency between forward and backward flows to determine areas subject to occlusion. That is, pixels can be occluded not just in one but in both inputs as shown in \Cref{fig:occtypes} and the more occlusion the more difficult it is to interpolate the affected area. After all, if an area is occluded in both inputs, the interpolation method needs to hallucinate the content.

Access to rich ground truth information such as the optical flow also allows us to analyze the $\text{PSNR}^{\ast}$ with respect to the motion magnitude, motion angle, as well as the photometric consistency. We will summarize some of our findings in \Cref{sec:results}. Such per-pixel evaluations prohibit the use of patch-wise metrics such as LPIPS~\cite{Zhang:2018:UED} though, which is a common error metric for frame interpolation. This is a clear limitation of our benchmark but we believe that it is a worthwhile trade-off that provides more insights.

\begin{figure*}\centering
    \setlength{\tabcolsep}{0.05cm}
    \setlength{\itemwidth}{2.41cm}
    \hspace*{-\tabcolsep}\begin{tabular}{ccccccc}
            \begin{tikzpicture}
                \node [anchor=south west, inner sep=0.0cm] (image) at (0,0) {
                    \includegraphics[width=\itemwidth]{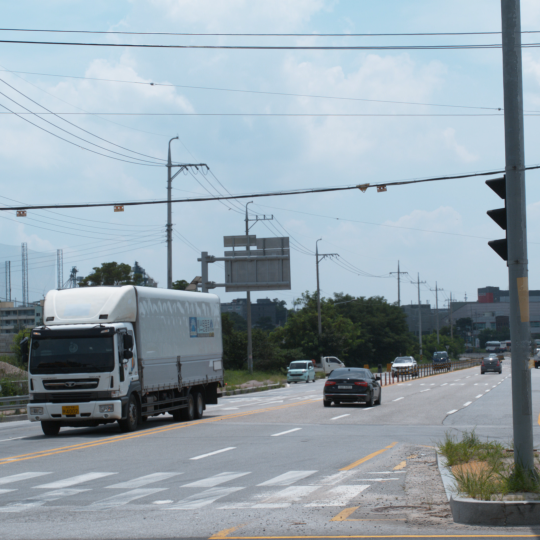}
                };
                \begin{scope}[x={(image.south east)},y={(image.north west)}]
                    \node [anchor=south, fill=white, inner sep=0.05cm] at (0.5,0.04) {\scalebox{0.75}{\tiny $\text{PSNR}^{\ast} = \text{37.27}, \text{PSNR}^{\ast}_{\sigma} = \text{28.47}$}};
                \end{scope}
            \end{tikzpicture}
        &
            \begin{tikzpicture}
                \node [anchor=south west, inner sep=0.0cm] (image) at (0,0) {
                    \includegraphics[width=\itemwidth]{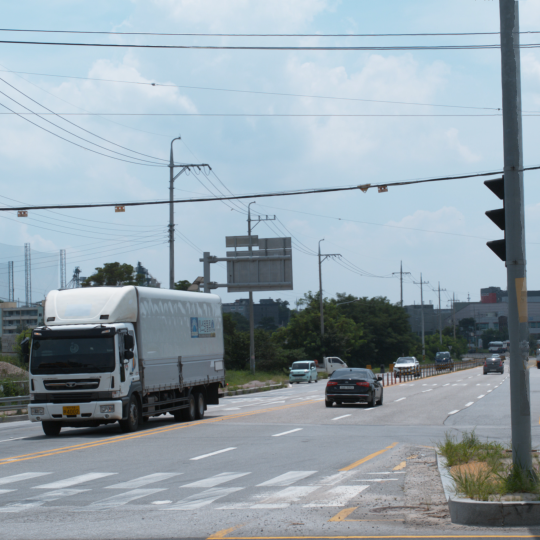}
                };
                \begin{scope}[x={(image.south east)},y={(image.north west)}]
                    \node [anchor=south, fill=white, inner sep=0.05cm] at (0.5,0.04) {\scalebox{0.75}{\tiny $\text{PSNR}^{\ast} = \text{34.59}, \text{PSNR}^{\ast}_{\sigma} = \text{26.13}$}};
                \end{scope}
            \end{tikzpicture}
        &
            \begin{tikzpicture}
                \node [anchor=south west, inner sep=0.0cm] (image) at (0,0) {
                    \includegraphics[width=\itemwidth]{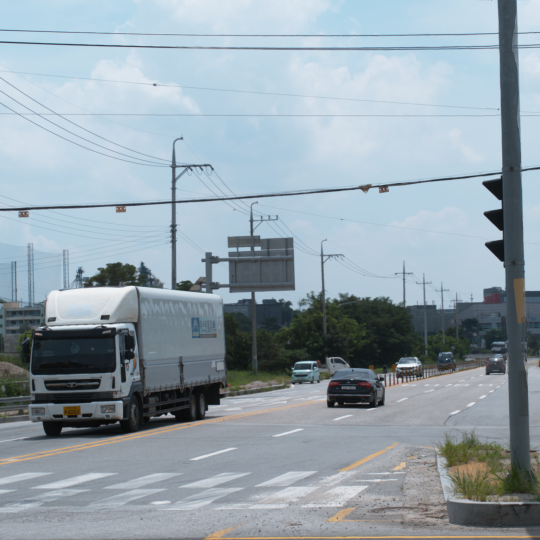}
                };
                \begin{scope}[x={(image.south east)},y={(image.north west)}]
                    \node [anchor=south, fill=white, inner sep=0.05cm] at (0.5,0.04) {\scalebox{0.75}{\tiny $\text{PSNR}^{\ast} = \text{34.49}, \text{PSNR}^{\ast}_{\sigma} = \text{25.99}$}};
                \end{scope}
            \end{tikzpicture}
        &
            \begin{tikzpicture}
                \node [anchor=south west, inner sep=0.0cm] (image) at (0,0) {
                    \includegraphics[width=\itemwidth]{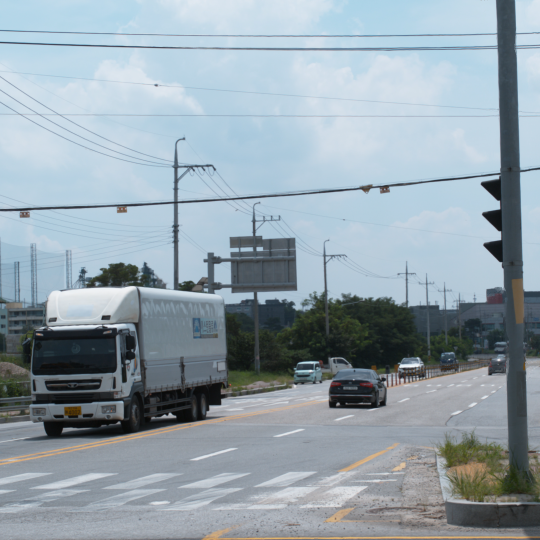}
                };
                \begin{scope}[x={(image.south east)},y={(image.north west)}]
                    \node [anchor=south, fill=white, inner sep=0.05cm] at (0.5,0.04) {\scalebox{0.75}{\tiny $\text{PSNR}^{\ast} = \text{35.26}, \text{PSNR}^{\ast}_{\sigma} = \text{27.59}$}};
                \end{scope}
            \end{tikzpicture}
        &
            \begin{tikzpicture}
                \node [anchor=south west, inner sep=0.0cm] (image) at (0,0) {
                    \includegraphics[width=\itemwidth]{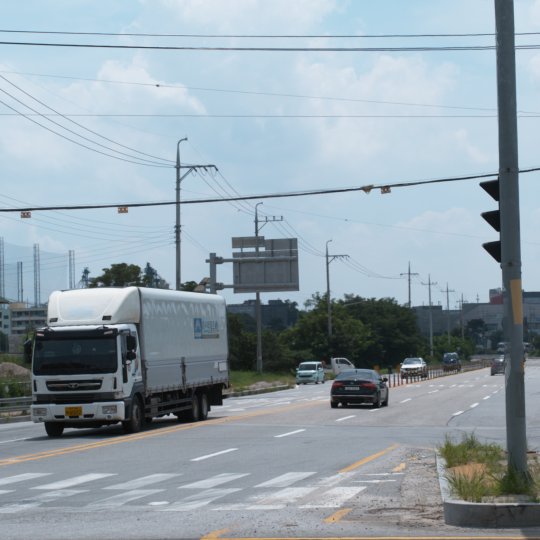}
                };
                \begin{scope}[x={(image.south east)},y={(image.north west)}]
                    \node [anchor=south, fill=white, inner sep=0.05cm] at (0.5,0.04) {\scalebox{0.75}{\tiny $\text{PSNR}^{\ast} = \text{35.87}, \text{PSNR}^{\ast}_{\sigma} = \text{27.52}$}};
                \end{scope}
            \end{tikzpicture}
        &
            \begin{tikzpicture}
                \node [anchor=south west, inner sep=0.0cm] (image) at (0,0) {
                    \includegraphics[width=\itemwidth]{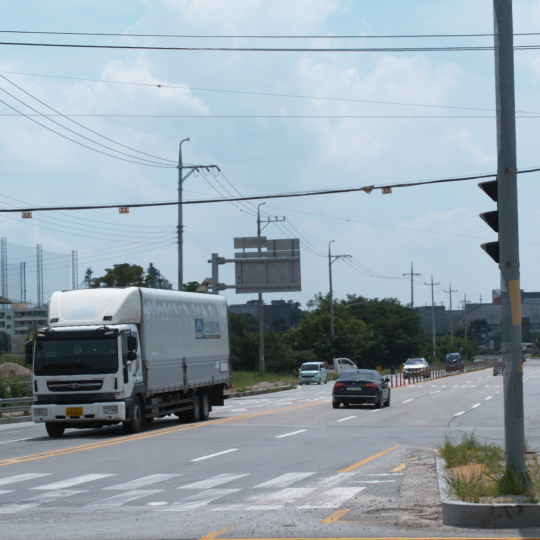}
                };
                \begin{scope}[x={(image.south east)},y={(image.north west)}]
                    \node [anchor=south, fill=white, inner sep=0.05cm] at (0.5,0.04) {\scalebox{0.75}{\tiny $\text{PSNR}^{\ast} = \text{36.53}, \text{PSNR}^{\ast}_{\sigma} = \text{27.11}$}};
                \end{scope}
            \end{tikzpicture}
        &
            \begin{tikzpicture}
                \node [anchor=south west, inner sep=0.0cm] (image) at (0,0) {
                    \includegraphics[width=\itemwidth]{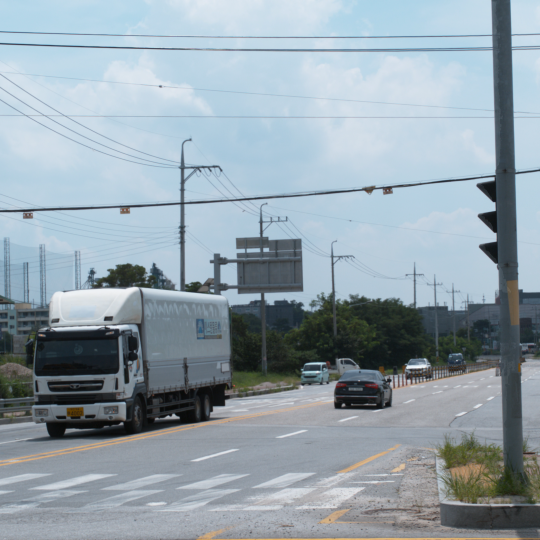}
                };
                \begin{scope}[x={(image.south east)},y={(image.north west)}]
                    \node [anchor=south, fill=white, inner sep=0.05cm] at (0.5,0.04) {\scalebox{0.75}{\tiny $\text{PSNR}^{\ast} = \text{38.63}, \text{PSNR}^{\ast}_{\sigma} = \text{27.24}$}};
                \end{scope}
            \end{tikzpicture}
        \\
            \begin{tikzpicture}
                \node [anchor=south west, inner sep=0.0cm] (image) at (0,0) {
                    \includegraphics[width=\itemwidth]{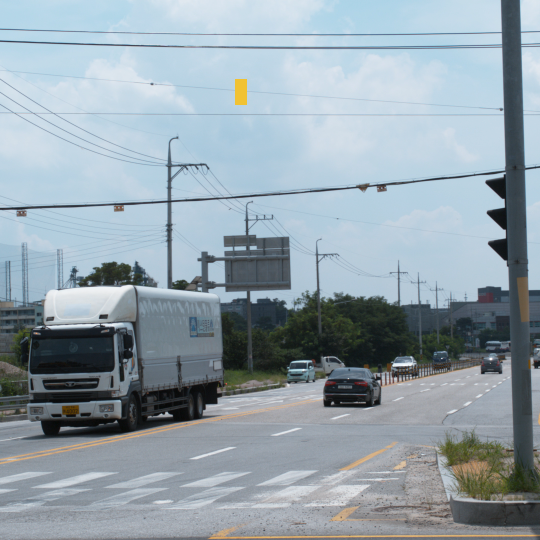}
                };
                \begin{scope}[x={(image.south east)},y={(image.north west)}]
                    \node [anchor=south, fill=white, inner sep=0.05cm] at (0.5,0.04) {\scalebox{0.75}{\tiny $\text{PSNR}^{\ast} = \text{37.27}, \text{PSNR}^{\ast}_{\sigma} = \text{20.42}$}};
                \end{scope}
            \end{tikzpicture}
        &
            \begin{tikzpicture}
                \node [anchor=south west, inner sep=0.0cm] (image) at (0,0) {
                    \includegraphics[width=\itemwidth]{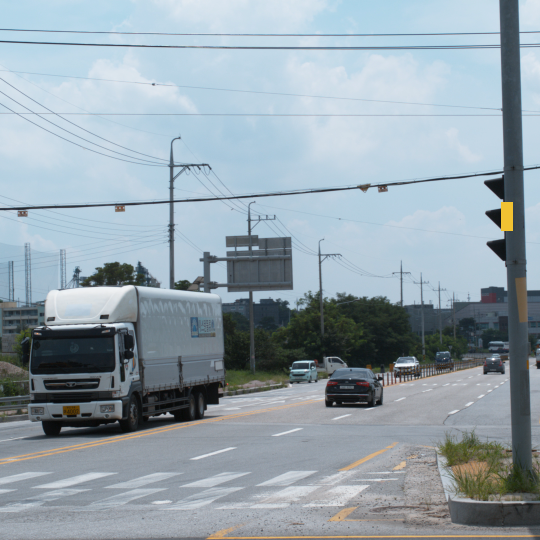}
                };
                \begin{scope}[x={(image.south east)},y={(image.north west)}]
                    \node [anchor=south, fill=white, inner sep=0.05cm] at (0.5,0.04) {\scalebox{0.75}{\tiny $\text{PSNR}^{\ast} = \text{34.59}, \text{PSNR}^{\ast}_{\sigma} = \text{18.81}$}};
                \end{scope}
            \end{tikzpicture}
        &
            \begin{tikzpicture}
                \node [anchor=south west, inner sep=0.0cm] (image) at (0,0) {
                    \includegraphics[width=\itemwidth]{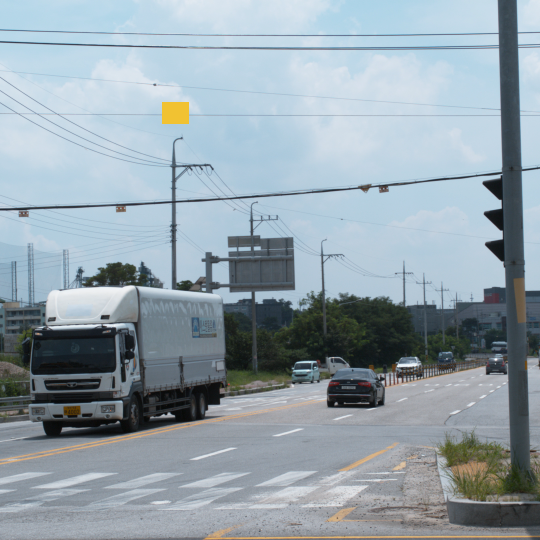}
                };
                \begin{scope}[x={(image.south east)},y={(image.north west)}]
                    \node [anchor=south, fill=white, inner sep=0.05cm] at (0.5,0.04) {\scalebox{0.75}{\tiny $\text{PSNR}^{\ast} = \text{34.49}, \text{PSNR}^{\ast}_{\sigma} = \text{19.10}$}};
                \end{scope}
            \end{tikzpicture}
        &
            \begin{tikzpicture}
                \node [anchor=south west, inner sep=0.0cm] (image) at (0,0) {
                    \includegraphics[width=\itemwidth]{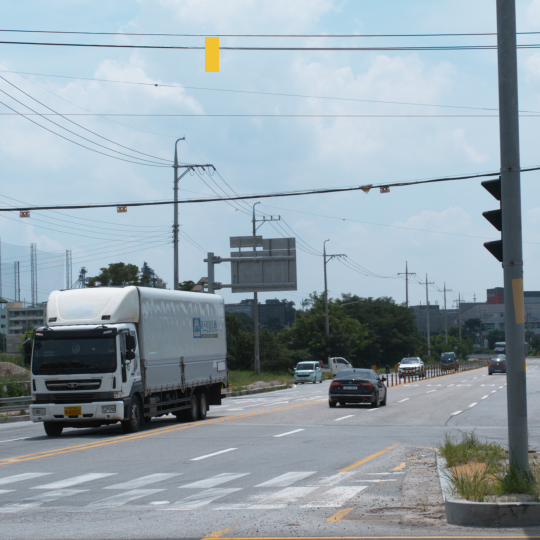}
                };
                \begin{scope}[x={(image.south east)},y={(image.north west)}]
                    \node [anchor=south, fill=white, inner sep=0.05cm] at (0.5,0.04) {\scalebox{0.75}{\tiny $\text{PSNR}^{\ast} = \text{35.26}, \text{PSNR}^{\ast}_{\sigma} = \text{19.36}$}};
                \end{scope}
            \end{tikzpicture}
        &
            \begin{tikzpicture}
                \node [anchor=south west, inner sep=0.0cm] (image) at (0,0) {
                    \includegraphics[width=\itemwidth]{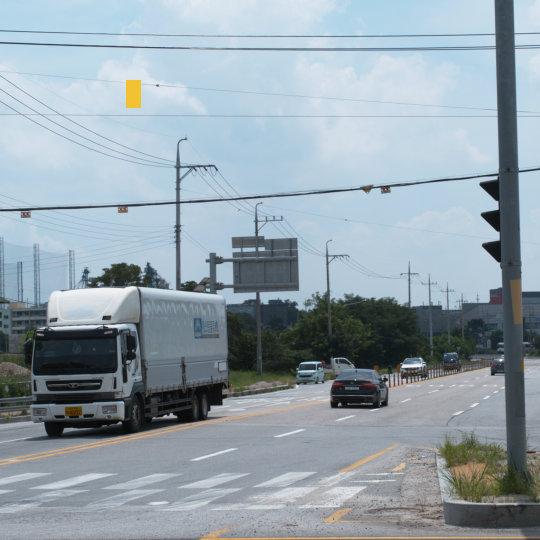}
                };
                \begin{scope}[x={(image.south east)},y={(image.north west)}]
                    \node [anchor=south, fill=white, inner sep=0.05cm] at (0.5,0.04) {\scalebox{0.75}{\tiny $\text{PSNR}^{\ast} = \text{35.87}, \text{PSNR}^{\ast}_{\sigma} = \text{19.50}$}};
                \end{scope}
            \end{tikzpicture}
        &
            \begin{tikzpicture}
                \node [anchor=south west, inner sep=0.0cm] (image) at (0,0) {
                    \includegraphics[width=\itemwidth]{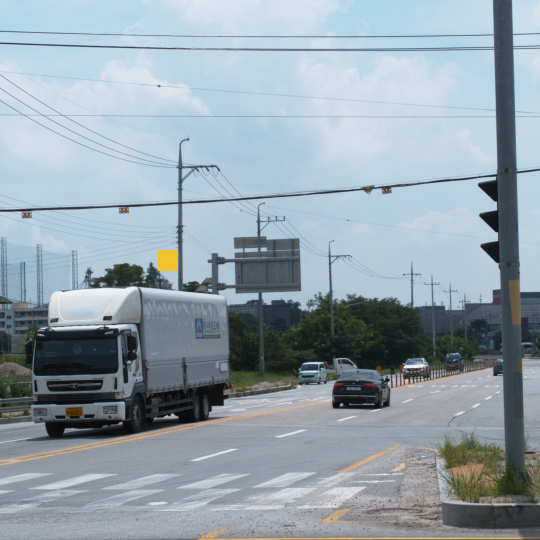}
                };
                \begin{scope}[x={(image.south east)},y={(image.north west)}]
                    \node [anchor=south, fill=white, inner sep=0.05cm] at (0.5,0.04) {\scalebox{0.75}{\tiny $\text{PSNR}^{\ast} = \text{36.53}, \text{PSNR}^{\ast}_{\sigma} = \text{20.32}$}};
                \end{scope}
            \end{tikzpicture}
        &
            \begin{tikzpicture}
                \node [anchor=south west, inner sep=0.0cm] (image) at (0,0) {
                    \includegraphics[width=\itemwidth]{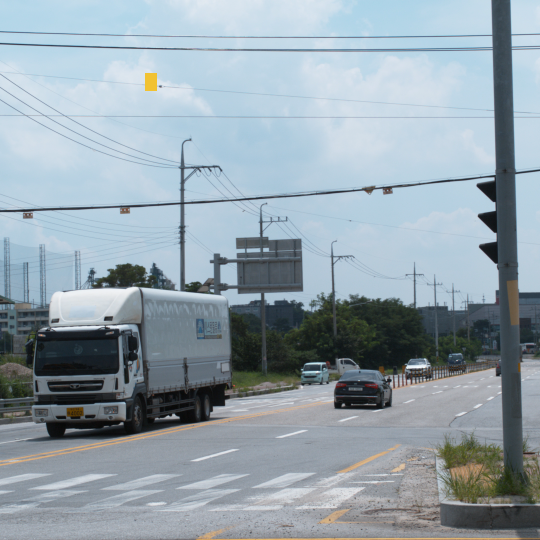}
                };
                \begin{scope}[x={(image.south east)},y={(image.north west)}]
                    \node [anchor=south, fill=white, inner sep=0.05cm] at (0.5,0.04) {\scalebox{0.75}{\tiny $\text{PSNR}^{\ast} = \text{38.63}, \text{PSNR}^{\ast}_{\sigma} = \text{20.82}$}};
                \end{scope}
            \end{tikzpicture}
        \\
            \footnotesize $t = 0.125$
        &
            \footnotesize $t = 0.25$
        &
            \footnotesize $t = 0.375$
        &
            \footnotesize $t = 0.5$
        &
            \footnotesize $t = 0.625$
        &
            \footnotesize $t = 0.75$
        &
            \footnotesize $t = 0.875$
        \\
    \end{tabular}\vspace{-0.2cm}
    \caption{To evaluate the temporal consistency when interpolating multiple in-between frames, it is common to analyze the per-timestep error. However, one can easily match the error of an interpolation result (first row) by perturbing the ground truth with random yellow rectangles (second row), which is inherently not consistent. As such, we do not solely rely on the $\text{PSNR}^{\ast}$ when performing multi-frame evaluations but focus on a newly proposed $\text{PSNR}^{\ast}_{\sigma}$ instead (notice the difference in $\text{PSNR}^{\ast}_{\sigma}$ between the first and second row).}\vspace{-0.2cm}
    \label{fig:mulframecstsy}
\end{figure*}

To evaluate the temporal consistency when interpolating multiple in-between frames, it is common to examine the per-timestep error. However, this is an inadequate measure since one can match these errors by perturbing the ground truth with random rectangles, which is inherently not temporally consistent. To address this we focus on the deviation when performing multi-frame evaluations:
\begin{equation}
    \text{PSNR}^{\ast}_{\sigma} = -10 \cdot \log_{10} \left( \sqrt{ \frac{1}{M - 1} \sum \left( \text{SE}_k - \mu \right)^{2} \thinspace } \right)
\end{equation}
which is akin to the $\text{PSNR}^{\ast}$ but we are taking the standard deviation of $\text{SE}_k$ instead of the mean. This metric better indicates temporal inconsistencies as shown in \Cref{fig:mulframecstsy}. Note that we use this $\text{PSNR}^{\ast}_{\sigma}$ only in the multi-frame evaluation in \Cref{tbl:mulframe} since all other experiments only evaluate one in-between frame where this new metric does not apply.

\begin{figure}\centering
    \setlength{\tabcolsep}{0.0cm}
    \renewcommand{\arraystretch}{1.2}
    \newcommand{\quantTit}[1]{\scriptsize #1}
    \newcommand{\quantSec}[1]{\scriptsize #1}
    \newcommand{\quantVal}[1]{\scalebox{0.83}[1.0]{$ #1 $}}
    \newcommand{\quantFirst}[1]{\usolid{\scalebox{0.83}[1.0]{$ #1 $}}}
    \newcommand{\quantSecond}[1]{\udotted{\scalebox{0.83}[1.0]{$ #1 $}}}
    \footnotesize
    \begin{tabularx}{\columnwidth}{@{\hspace{0.1cm}} X P{1.5cm} P{1.5cm} P{1.5cm} P{1.5cm}}
        \toprule
             & \quantTit{Vimeo-90k} & \quantTit{Xiph - 448} & \quantTit{X-TEST - 448} & \quantTit{Ours - 448}
        \\ \midrule
Vimeo-90k & \quantVal{-} & \quantVal{0.830} & \quantVal{0.042} & \quantVal{0.842}
\\
Xiph - 448 & \quantVal{0.830} & \quantVal{-} & \quantVal{0.152} & \quantVal{0.770}
\\
X-TEST - 448 & \quantVal{0.042} & \quantVal{0.152} & \quantVal{-} & \quantVal{0.236}
\\
Ours - 448 & \quantVal{0.842} & \quantVal{0.770} & \quantVal{0.236} & \quantVal{-}
        \\ \bottomrule
    \end{tabularx}\vspace{-0.2cm}
    \captionof{table}{Analyzing the Spearman correlation from the evaluation in \Cref{tbl:commonrank}. Notice that the results on Vimeo-90k, Xiph, and our benchmark correlate with each other but X-TEST does not.}\vspace{-0.1cm}
    \label{tbl:commoncorr}
\end{figure}

\begin{figure*}\centering
    \setlength{\tabcolsep}{0.0cm}
    \renewcommand{\arraystretch}{1.2}
    \newcommand{\quantTit}[1]{\multicolumn{5}{c}{\scriptsize #1}}
    \newcommand{\quantSec}[1]{\scriptsize #1}
    \newcommand{\quantVal}[1]{\scalebox{0.83}[1.0]{$ #1 $}}
    \newcommand{\quantFirst}[1]{\usolid{\scalebox{0.83}[1.0]{$ #1 $}}}
    \newcommand{\quantSecond}[1]{\udotted{\scalebox{0.83}[1.0]{$ #1 $}}}
    \footnotesize
    \begin{tabularx}{\textwidth}{@{\hspace{0.1cm}} X P{1.3cm} @{\hspace{-0.3cm}} P{1.2cm} @{\hspace{-0.3cm}} P{1.2cm} @{\hspace{-0.3cm}} P{1.2cm} @{\hspace{-0.3cm}} P{1.2cm} P{1.3cm} @{\hspace{-0.3cm}} P{1.2cm} @{\hspace{-0.3cm}} P{1.2cm} @{\hspace{-0.3cm}} P{1.2cm} @{\hspace{-0.3cm}} P{1.2cm} P{1.3cm} @{\hspace{-0.3cm}} P{1.2cm} @{\hspace{-0.3cm}} P{1.2cm} @{\hspace{-0.3cm}} P{1.2cm} @{\hspace{-0.3cm}} P{1.2cm}}
        \toprule
             & \quantTit{1K} & \quantTit{2K} & \quantTit{4K}
        \\ \cmidrule(l{2pt}r{2pt}){2-6} \cmidrule(l{2pt}r{2pt}){7-11} \cmidrule(l{2pt}r{2pt}){12-16}
             & \quantSec{rank} & \quantSec{all} & \quantSec{0-occ.} & \quantSec{1-occ.} & \quantSec{2-occ.} & \quantSec{rank} & \quantSec{all} & \quantSec{0-occ.} & \quantSec{1-occ.} & \quantSec{2-occ.} & \quantSec{rank} & \quantSec{all} & \quantSec{0-occ.} & \quantSec{1-occ.} & \quantSec{2-occ.}
        \\ \midrule
ABME~\cite{Park:2021:ABM} & \quantVal{\hphantom{0}2\text{ / } 10} & \quantVal{28.63} & \quantVal{31.62} & \quantVal{21.36} & \quantVal{16.95} & \quantVal{\hphantom{0}5\text{ / } 10} & \quantVal{23.95} & \quantVal{25.35} & \quantVal{18.41} & \quantVal{15.54} & \quantVal{\hphantom{0}8\text{ / } 10} & \quantVal{18.35} & \quantVal{18.80} & \quantVal{15.41} & \quantVal{13.97}
\\
AMT-G~\cite{Li:2023:AMT} & \quantVal{\hphantom{0}7\text{ / } 10} & \quantVal{28.18} & \quantVal{30.62} & \quantVal{21.35} & \quantFirst{17.80} & \quantVal{10\text{ / } 10} & \quantVal{20.97} & \quantVal{21.71} & \quantVal{17.00} & \quantVal{15.72} & \quantVal{-} & \quantVal{-} & \quantVal{-} & \quantVal{-} & \quantVal{-}
\\
CtxSyn~\cite{Niklaus:2018:CAS} & \quantVal{\hphantom{0}1\text{ / } 10} & \quantFirst{29.36} & \quantFirst{32.72} & \quantVal{21.81} & \quantVal{17.34} & \quantVal{\hphantom{0}7\text{ / } 10} & \quantVal{23.33} & \quantVal{24.65} & \quantVal{17.91} & \quantVal{15.81} & \quantVal{\hphantom{0}7\text{ / } 10} & \quantVal{18.80} & \quantVal{19.47} & \quantVal{14.98} & \quantVal{13.75}
\\
DAIN~\cite{Bao:2019:DAV} & \quantVal{\hphantom{0}8\text{ / } 10} & \quantVal{27.83} & \quantVal{31.47} & \quantVal{19.97} & \quantVal{16.94} & \quantVal{\hphantom{0}3\text{ / } 10} & \quantVal{26.81} & \quantVal{29.77} & \quantVal{19.32} & \quantVal{16.40} & \quantVal{-} & \quantVal{-} & \quantVal{-} & \quantVal{-} & \quantVal{-}
\\
FLDR~\cite{Nottebaum:2022:EFE} & \quantVal{\hphantom{0}9\text{ / } 10} & \quantVal{24.09} & \quantVal{25.80} & \quantVal{18.14} & \quantVal{15.56} & \quantVal{\hphantom{0}6\text{ / } 10} & \quantVal{23.80} & \quantVal{25.38} & \quantVal{17.96} & \quantVal{15.12} & \quantVal{\hphantom{0}3\text{ / } 10} & \quantVal{23.53} & \quantVal{25.05} & \quantVal{17.68} & \quantVal{14.47}
\\
M2M~\cite{Hu:2022:MMS} & \quantVal{\hphantom{0}3\text{ / } 10} & \quantVal{28.61} & \quantVal{31.73} & \quantVal{21.21} & \quantVal{17.16} & \quantVal{\hphantom{0}4\text{ / } 10} & \quantVal{25.81} & \quantVal{28.12} & \quantVal{18.95} & \quantVal{15.99} & \quantVal{\hphantom{0}5\text{ / } 10} & \quantVal{20.61} & \quantVal{21.44} & \quantVal{16.29} & \quantVal{14.49}
\\
SoftSplat~\cite{Niklaus:2020:SSV} & \quantVal{\hphantom{0}4\text{ / } 10} & \quantVal{28.55} & \quantVal{31.28} & \quantVal{21.46} & \quantVal{17.65} & \quantVal{\hphantom{0}9\text{ / } 10} & \quantVal{22.59} & \quantVal{23.58} & \quantVal{17.91} & \quantVal{16.05} & \quantVal{\hphantom{0}6\text{ / } 10} & \quantVal{18.85} & \quantVal{19.45} & \quantVal{15.28} & \quantVal{14.04}
\\
SplatSyn~\cite{Niklaus:2023:SSV} & \quantVal{\hphantom{0}5\text{ / } 10} & \quantVal{28.45} & \quantVal{30.89} & \quantVal{21.71} & \quantVal{17.20} & \quantVal{\hphantom{0}1\text{ / } 10} & \quantFirst{28.40} & \quantFirst{30.91} & \quantFirst{21.43} & \quantFirst{16.71} & \quantVal{\hphantom{0}1\text{ / } 10} & \quantFirst{27.40} & \quantFirst{29.71} & \quantFirst{20.48} & \quantFirst{15.97}
\\
UPR-Net-L~\cite{Jin:2023:UPR} & \quantVal{\hphantom{0}6\text{ / } 10} & \quantVal{28.31} & \quantVal{30.28} & \quantFirst{22.14} & \quantVal{17.46} & \quantVal{\hphantom{0}2\text{ / } 10} & \quantVal{27.09} & \quantVal{28.76} & \quantVal{21.20} & \quantVal{16.47} & \quantVal{\hphantom{0}2\text{ / } 10} & \quantVal{25.69} & \quantVal{27.05} & \quantVal{20.28} & \quantVal{14.65}
\\
XVFI~\cite{Sim:2021:XVF} & \quantVal{10\text{ / } 10} & \quantVal{23.73} & \quantVal{25.36} & \quantVal{17.90} & \quantVal{15.33} & \quantVal{\hphantom{0}8\text{ / } 10} & \quantVal{23.25} & \quantVal{24.74} & \quantVal{17.54} & \quantVal{14.95} & \quantVal{\hphantom{0}4\text{ / } 10} & \quantVal{22.48} & \quantVal{23.71} & \quantVal{17.15} & \quantVal{14.33}
        \\ \bottomrule
    \end{tabularx}\vspace{-0.2cm}
    \captionof{table}{Results of representative interpolation methods on our benchmark across multiple resolutions. We report $\text{PSNR}^{\ast}$ separately across all areas (all), non-occluded areas (0-occ.), areas that are occluded in one input (1-occ.), and areas that are occluded in both inputs (2-occ.). We are unfortunately unable to report some numbers denoted with a dash due to not having enough GPU memory to run these methods.}\vspace{-0.3cm}
    \label{tbl:main}
\end{figure*}

\section{Submission Page}
\label{sec:page}

We follow the common paradigm of benchmarks in other areas~\cite{Baker:2011:DBE,Geiger:2012:AWR, Butler:2012:NOS} and provide a submission page. This ensures consistent and comparable results across papers through a fixed evaluation procedure. Furthermore, a common leaderboard provides easy discoverability of the current state of the art, even in case of multiple concurrent developments.

We provide a Python library to simplify the submission process, which uploads compressed quantized interpolation results. To be mindful of how much data the participants of our benchmark need to upload, we perform the multi-frame evaluation only at a 1K resolution. Also, we would like to save researcher's time when writing papers and provide features that export the LaTeX representation of many of the tables and figures shown in this paper. Another feature is an anomaly detector, which detects accidental mistakes as well as intentional cheating attempts. For example and as discussed in \Cref{sec:limitations}, performing ensembling can significantly alter the outcome which is why submissions have to indicate whether or not they have used ensembling and we have measures in place that detect false reporting.

Lastly, should submissions choose to include their inference code, we will run them on a reference machine to measure and report the computational efficiency. Measuring the computational efficiency correctly is tricky, since it requires knowledge of the asynchronous nature of GPU compute as well as standardized test hardware. To reduce the burden on researchers, we hence provide such data ourselves. Note that we refrain from utilizing features such as PyTorch compilations or TensorRT, but this may change in the future.

\section{Results}
\label{sec:results}

We evaluate 21 frame interpolation methods on our benchmark, covering a wide spectrum of the existing state-of-the-art literature. Due to space constraints however, we only include 10 representative methods in the main paper and kindly refer to the supplementary for the full picture.

\begin{figure*}\centering
    \setlength{\tabcolsep}{0.05cm}
    \setlength{\itemwidth}{3.38cm}
    \hspace*{-\tabcolsep}\begin{tabular}{P{\itemwidth} P{\itemwidth} P{\itemwidth} P{\itemwidth} P{\itemwidth}}
            \hspace{-0.15cm}\includegraphics[]{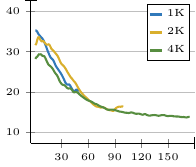}
        &
            \hspace{-0.15cm}\includegraphics[]{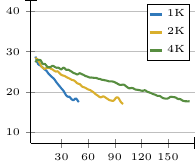}
        &
            \hspace{-0.15cm}\includegraphics[]{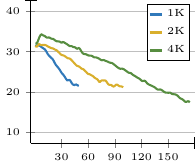}
        &
            \hspace{-0.15cm}\includegraphics[]{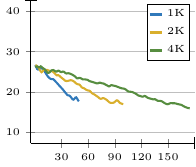}
        &
            \hspace{-0.15cm}\includegraphics[]{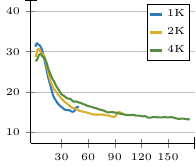}
        \\
            \footnotesize ABME~\cite{Park:2021:ABM}
        &
            \footnotesize FLDR~\cite{Nottebaum:2022:EFE}
        &
            \footnotesize SplatSyn~\cite{Niklaus:2023:SSV}
        &
            \footnotesize XVFI~\cite{Sim:2021:XVF}
        &
            \footnotesize XVFI$_v$~\cite{Sim:2021:XVF}
        \\
    \end{tabular}\vspace{-0.2cm}
    \caption{Analyzing the per-pixel interpolation quality (Y axis, which is the $\text{PSNR}^{\ast}$) with respect to the motion magnitude (X axis, which is the motion in pixels). Notice that multi-scale methods like FLDR and SplatSyn better handle larger motions if the resolution is higher.}\vspace{-0.2cm}
    \label{fig:motionmag}
\end{figure*}

\begin{figure*}\centering
    \setlength{\tabcolsep}{0.05cm}
    \setlength{\itemwidth}{3.38cm}
    \hspace*{-\tabcolsep}\begin{tabular}{P{\itemwidth} P{\itemwidth} P{\itemwidth} P{\itemwidth} P{\itemwidth}}
            \includegraphics[]{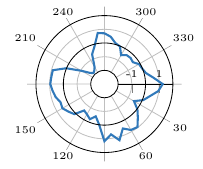}
        &
            \includegraphics[]{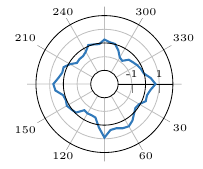}
        &
            \includegraphics[]{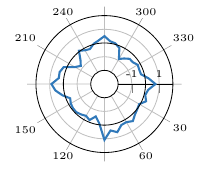}
        &
            \includegraphics[]{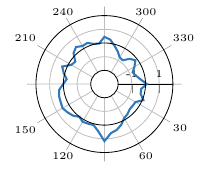}
        &
            \includegraphics[]{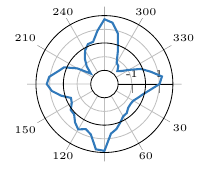}
        \\
            \footnotesize ABME~\cite{Park:2021:ABM}
        &
            \footnotesize FLDR~\cite{Nottebaum:2022:EFE}
        &
            \footnotesize SplatSyn~\cite{Niklaus:2023:SSV}
        &
            \footnotesize XVFI~\cite{Sim:2021:XVF}
        &
            \footnotesize XVFI$_v$~\cite{Sim:2021:XVF}
        \\
    \end{tabular}\vspace{-0.2cm}
    \caption{Analyzing the per-pixel interpolation quality with respect to the motion angle at a 1K resolution where the plot indicates the difference between the overall $\text{PSNR}^{\ast}$ and the $\text{PSNR}^{\ast}$ in each direction. Notice the varying bias towards axis-aligned motion across all depicted methods, which manifests in a rhombus-like shape of the plotted per-angle deviation of the interpolation quality.}\vspace{-0.3cm}
    \label{fig:motionangle}
\end{figure*}

Perhaps the most prominent outcome of our proposed benchmark is the quantitative evaluation across multiple resolutions in \Cref{tbl:main}, where we not only analyze the overall $\text{PSNR}^{\ast}$ but also the interpolation quality in regions subject to increasing amounts of occlusion. Unsurprisingly, the larger the resolution and hence motion or the more occlusion the lower the $\text{PSNR}^{\ast}$ becomes. Surprisingly, the amount greatly differs and some methods like FLDR only degrade by $0.56$ dB going from 1K to 4K whereas others like ABME degrade significantly by $10.28$ dB instead.

To delve deeper into this phenomenon, we analyze the interpolation quality with respect to the motion magnitude in \Cref{fig:motionmag}. We find that multi-scale methods like FLDR and SplatSyn better handle larger motions if the input resolution is higher, and we hypothesize that this is due to their coarse-to-fine nature benefiting from using more scales when given larger inputs. Another interesting observation is that the training data significantly impacts the ability to handle large motion. Notice the difference between XVFI and XVFI$_v$, which are the same but the former was trained on X-TRAIN while the latter was instead trained on Vimeo-90k which is much smaller in resolution.

Moving on to an evaluation of the interpolation quality with respect to the motion angle in \Cref{fig:motionangle}, we find that all methods are biased towards axis-aligned motion which manifests in a rhombus-like shape of the plots. This comes as no surprise since \Cref{fig:histangle} has demonstrated that usual datasets are subject to a bias towards horizontal motion, and the augmentation pipeline when training a frame interpolation network typically includes random rotations by ninety degrees which extends the horizontal bias to both axis.

Our benchmark also makes it possible to evaluate the interpolation quality with respect to the photometric consistency. We do not have any interesting findings to report on this matter though, which is why we will omit them in our main paper for brevity. Nevertheless, we include this evaluation in our benchmark since it enables future research that intends to better handle photometric inconsistencies.

\begin{figure*}\centering
    \setlength{\tabcolsep}{0.0cm}
    \renewcommand{\arraystretch}{1.2}
    \newcommand{\quantTit}[1]{\multicolumn{8}{c}{\scriptsize #1}}
    \newcommand{\quantSec}[1]{\scriptsize \scalebox{0.78}[1.0]{$ #1 $}}
    \newcommand{\quantVal}[1]{\scalebox{0.83}[1.0]{$ #1 $}}
    \newcommand{\quantFirst}[1]{\usolid{\scalebox{0.83}[1.0]{$ #1 $}}}
    \newcommand{\quantSecond}[1]{\udotted{\scalebox{0.83}[1.0]{$ #1 $}}}
    \footnotesize
    \begin{tabularx}{\textwidth}{@{\hspace{0.1cm}} X P{1.3cm} @{\hspace{-0.3cm}} P{1.17cm} @{\hspace{-0.3cm}} P{1.17cm} @{\hspace{-0.3cm}} P{1.17cm} @{\hspace{-0.3cm}} P{1.17cm} @{\hspace{-0.3cm}} P{1.17cm} @{\hspace{-0.3cm}} P{1.17cm} @{\hspace{-0.3cm}} P{1.17cm} P{1.3cm} @{\hspace{-0.3cm}} P{1.17cm} @{\hspace{-0.3cm}} P{1.17cm} @{\hspace{-0.3cm}} P{1.17cm} @{\hspace{-0.3cm}} P{1.17cm} @{\hspace{-0.3cm}} P{1.17cm} @{\hspace{-0.3cm}} P{1.17cm} @{\hspace{-0.3cm}} P{1.17cm}}
        \toprule
             & \quantTit{$\text{PSNR}^{\ast}$} & \quantTit{$\text{PSNR}^{\ast}_{\sigma}$}
        \\ \cmidrule(l{2pt}r{2pt}){2-9} \cmidrule(l{2pt}r{2pt}){10-17} 
             & \scriptsize rank & \quantSec{$t = 0.125$} & \quantSec{$t = 0.25$} & \quantSec{$t = 0.375$} & \quantSec{$t = 0.5$} & \quantSec{$t = 0.625$} & \quantSec{$t = 0.75$} & \quantSec{$t = 0.875$} & \scriptsize rank & \quantSec{$t = 0.125$} & \quantSec{$t = 0.25$} & \quantSec{$t = 0.375$} & \quantSec{$t = 0.5$} & \quantSec{$t = 0.625$} & \quantSec{$t = 0.75$} & \quantSec{$t = 0.875$}
        \\ \midrule
ABME~\cite{Park:2021:ABM} & \quantVal{\hphantom{0}3\text{ / } 10} & \quantVal{31.00} & \quantVal{29.43} & \quantVal{28.99} & \quantVal{28.63} & \quantVal{28.67} & \quantVal{28.91} & \quantVal{30.57} & \quantVal{\hphantom{0}3\text{ / } 10} & \quantVal{21.38} & \quantVal{20.26} & \quantVal{20.05} & \quantVal{19.72} & \quantVal{19.86} & \quantVal{19.95} & \quantVal{21.02}
\\
AMT-G~\cite{Li:2023:AMT} & \quantVal{10\text{ / } 10} & \quantVal{16.69} & \quantVal{17.68} & \quantVal{19.44} & \quantVal{28.18} & \quantVal{19.47} & \quantVal{17.37} & \quantVal{16.29} & \quantVal{10\text{ / } 10} & \quantVal{12.14} & \quantVal{12.74} & \quantVal{13.86} & \quantVal{19.28} & \quantVal{13.88} & \quantVal{12.50} & \quantVal{11.86}
\\
CtxSyn~\cite{Niklaus:2018:CAS} & \quantVal{\hphantom{0}5\text{ / } 10} & \quantVal{28.72} & \quantVal{29.07} & \quantFirst{29.30} & \quantFirst{29.36} & \quantFirst{29.26} & \quantVal{29.14} & \quantVal{29.11} & \quantVal{\hphantom{0}1\text{ / } 10} & \quantVal{21.87} & \quantFirst{21.19} & \quantFirst{20.77} & \quantFirst{20.55} & \quantFirst{20.56} & \quantFirst{20.85} & \quantFirst{21.61}
\\
DAIN~\cite{Bao:2019:DAV} & \quantVal{\hphantom{0}7\text{ / } 10} & \quantVal{27.14} & \quantVal{26.81} & \quantVal{27.15} & \quantVal{27.83} & \quantVal{27.05} & \quantVal{26.88} & \quantVal{27.55} & \quantVal{\hphantom{0}7\text{ / } 10} & \quantVal{19.51} & \quantVal{18.68} & \quantVal{18.59} & \quantVal{18.71} & \quantVal{18.55} & \quantVal{18.73} & \quantVal{19.80}
\\
FLDR~\cite{Nottebaum:2022:EFE} & \quantVal{\hphantom{0}8\text{ / } 10} & \quantVal{27.30} & \quantVal{25.24} & \quantVal{24.38} & \quantVal{24.09} & \quantVal{24.26} & \quantVal{24.97} & \quantVal{26.82} & \quantVal{\hphantom{0}8\text{ / } 10} & \quantVal{19.17} & \quantVal{17.84} & \quantVal{17.37} & \quantVal{17.20} & \quantVal{17.27} & \quantVal{17.68} & \quantVal{18.92}
\\
M2M~\cite{Hu:2022:MMS} & \quantVal{\hphantom{0}1\text{ / } 10} & \quantFirst{31.83} & \quantFirst{29.86} & \quantVal{28.93} & \quantVal{28.61} & \quantVal{28.57} & \quantVal{29.19} & \quantVal{30.89} & \quantVal{\hphantom{0}4\text{ / } 10} & \quantVal{21.94} & \quantVal{20.53} & \quantVal{19.87} & \quantVal{19.52} & \quantVal{19.48} & \quantVal{19.81} & \quantVal{20.99}
\\
SoftSplat~\cite{Niklaus:2020:SSV} & \quantVal{\hphantom{0}4\text{ / } 10} & \quantVal{30.78} & \quantVal{29.30} & \quantVal{28.79} & \quantVal{28.55} & \quantVal{28.52} & \quantVal{28.78} & \quantVal{30.04} & \quantVal{\hphantom{0}5\text{ / } 10} & \quantVal{21.19} & \quantVal{20.19} & \quantVal{19.82} & \quantVal{19.62} & \quantVal{19.59} & \quantVal{19.73} & \quantVal{20.53}
\\
SplatSyn~\cite{Niklaus:2023:SSV} & \quantVal{\hphantom{0}2\text{ / } 10} & \quantVal{31.78} & \quantVal{29.72} & \quantVal{28.70} & \quantVal{28.45} & \quantVal{28.53} & \quantFirst{29.30} & \quantFirst{31.02} & \quantVal{\hphantom{0}2\text{ / } 10} & \quantFirst{22.32} & \quantVal{20.87} & \quantVal{20.23} & \quantVal{19.97} & \quantVal{19.96} & \quantVal{20.29} & \quantVal{21.50}
\\
UPR-Net-L~\cite{Jin:2023:UPR} & \quantVal{\hphantom{0}6\text{ / } 10} & \quantVal{27.90} & \quantVal{27.85} & \quantVal{28.34} & \quantVal{28.31} & \quantVal{28.03} & \quantVal{27.69} & \quantVal{27.90} & \quantVal{\hphantom{0}6\text{ / } 10} & \quantVal{19.06} & \quantVal{19.38} & \quantVal{19.91} & \quantVal{19.77} & \quantVal{19.69} & \quantVal{19.46} & \quantVal{19.70}
\\
XVFI~\cite{Sim:2021:XVF} & \quantVal{\hphantom{0}9\text{ / } 10} & \quantVal{26.43} & \quantVal{24.71} & \quantVal{23.99} & \quantVal{23.73} & \quantVal{23.83} & \quantVal{24.39} & \quantVal{25.85} & \quantVal{\hphantom{0}9\text{ / } 10} & \quantVal{18.54} & \quantVal{17.39} & \quantVal{16.95} & \quantVal{16.79} & \quantVal{16.85} & \quantVal{17.20} & \quantVal{18.16}
        \\ \bottomrule
    \end{tabularx}\vspace{-0.2cm}
    \captionof{table}{Evaluation of the ability to perform multi-frame interpolation based on seven in-between ground truth frames at a 1K resolution.}\vspace{-0.2cm}
    \label{tbl:mulframe}
\end{figure*}

Typical applications of frame interpolation are interested in more than just a single interpolation result that is temporally centered between the inputs. To assess this property, we perform a multi-frame evaluation based on seven in-between ground truth frames at a 1K resolution in \Cref{tbl:mulframe}. The findings of this experiment are mostly unremarkable, all methods appear temporally consistent. The only outlier seems to be DAIN, which exhibits its lowest $\text{PSNR}^{\ast}$ at $t = \left\{ 0.25, 0.75 \right\}$. However, DAIN is visually temporally consistent and the $\text{PSNR}^{\ast}_{\sigma}$ is not subject to this dip either.

Lastly, we analyze the computational efficiency during inference in \Cref{tbl:timetable}. Specifically, we perform single-frame and multi-frame interpolation at various resolutions and measure the runtime on a single A100 (the exact environment is outlined in the supplementary). Note that the reported numbers can be considered the worst-case scenario for practical applications since we do not do batching. In real-world applications, one should be able to achieve some performance gains due to batched parallelism.

\begin{figure}\centering
    \setlength{\tabcolsep}{0.0cm}
    \renewcommand{\arraystretch}{1.2}
    \newcommand{\quantTit}[1]{\multicolumn{3}{c}{\scriptsize #1}}
    \newcommand{\quantSec}[1]{\scriptsize #1}
    \newcommand{\quantInd}[1]{\scriptsize #1}
    \newcommand{\quantVal}[1]{\scalebox{0.83}[1.0]{$ #1 $}}
    \newcommand{\quantFirst}[1]{\usolid{\scalebox{0.83}[1.0]{$ #1 $}}}
    \newcommand{\quantSecond}[1]{\udotted{\scalebox{0.83}[1.0]{$ #1 $}}}
    \footnotesize
    \begin{tabularx}{\columnwidth}{@{\hspace{0.1cm}} X P{1.05cm} @{\hspace{-0.1cm}} P{1.05cm} @{\hspace{-0.1cm}} P{1.05cm} P{1.05cm} @{\hspace{-0.1cm}} P{1.05cm} @{\hspace{-0.1cm}} P{1.05cm}}
        \toprule
            & \quantTit{1 frame} & \quantTit{7 frames}
        \\ \cmidrule(l{2pt}r{2pt}){2-4} \cmidrule(l{2pt}r{2pt}){5-7}
            & {\scriptsize 1K} & {\scriptsize 2K} & {\scriptsize 4K} & {\scriptsize 1K} & {\scriptsize 2K} & {\scriptsize 4K}
        \\ \midrule
ABME~\cite{Park:2021:ABM} & \quantVal{0.29} & \quantVal{1.08} & \quantVal{4.34} & \quantVal{\hphantom{0}}\quantVal{2.00} & \quantVal{\hphantom{0}}\quantVal{7.88} & \quantVal{30.51}
\\
AMT-G~\cite{Li:2023:AMT} & \quantVal{0.07} & \quantVal{0.31} & \quantVal{-} & \quantVal{\hphantom{0}}\quantVal{0.48} & \quantVal{\hphantom{0}}\quantVal{2.20} & \quantVal{-}
\\
CtxSyn~\cite{Niklaus:2018:CAS} & \quantVal{0.06} & \quantVal{0.19} & \quantVal{0.76} & \quantVal{\hphantom{0}}\quantVal{0.23} & \quantVal{\hphantom{0}}\quantVal{0.86} & \quantVal{\hphantom{0}}\quantVal{3.49}
\\
DAIN~\cite{Bao:2019:DAV} & \quantVal{0.21} & \quantVal{0.81} & \quantVal{-} & \quantVal{\hphantom{0}}\quantVal{0.68} & \quantVal{\hphantom{0}}\quantVal{2.87} & \quantVal{-}
\\
FLDR~\cite{Nottebaum:2022:EFE} & \quantVal{0.05} & \quantVal{0.14} & \quantVal{0.59} & \quantVal{\hphantom{0}}\quantVal{0.34} & \quantVal{\hphantom{0}}\quantVal{0.97} & \quantVal{\hphantom{0}}\quantVal{4.00}
\\
M2M~\cite{Hu:2022:MMS} & \quantFirst{0.02} & \quantVal{0.05} & \quantVal{0.17} & \quantVal{\hphantom{0}}\quantVal{0.17} & \quantVal{\hphantom{0}}\quantVal{0.34} & \quantVal{\hphantom{0}}\quantVal{1.18}
\\
SoftSplat~\cite{Niklaus:2020:SSV} & \quantVal{0.06} & \quantVal{0.21} & \quantVal{0.81} & \quantVal{\hphantom{0}}\quantVal{0.30} & \quantVal{\hphantom{0}}\quantVal{1.08} & \quantVal{\hphantom{0}}\quantVal{4.23}
\\
SplatSyn~\cite{Niklaus:2023:SSV} & \quantVal{0.03} & \quantFirst{0.04} & \quantFirst{0.09} & \quantVal{\hphantom{0}}\quantFirst{0.04} & \quantVal{\hphantom{0}}\quantFirst{0.06} & \quantVal{\hphantom{0}}\quantFirst{0.20}
\\
UPR-Net-L~\cite{Jin:2023:UPR} & \quantVal{0.06} & \quantVal{0.16} & \quantVal{0.61} & \quantVal{\hphantom{0}}\quantVal{0.41} & \quantVal{\hphantom{0}}\quantVal{1.17} & \quantVal{\hphantom{0}}\quantVal{4.30}
\\
XVFI~\cite{Sim:2021:XVF} & \quantVal{0.05} & \quantVal{0.12} & \quantVal{0.61} & \quantVal{\hphantom{0}}\quantVal{0.34} & \quantVal{\hphantom{0}}\quantVal{0.90} & \quantVal{\hphantom{0}}\quantVal{4.37}
        \\ \bottomrule
    \end{tabularx}\vspace{-0.2cm}
    \captionof{table}{Overview of the runtime during inference in seconds. The missing numbers are due to out-of-memory issues.}\vspace{-0.3cm}
    \label{tbl:timetable}
\end{figure}

\section{Limitations}
\label{sec:limitations}

Creating a benchmark requires making many difficult decisions. Each of these decisions is well motivated but often incurs a dept, such as trading the ability to have access to rich ground truth information through synthetic data with a potential domain gap (which in our case does not seem to be present as discussed in \Cref{sec:dataset}). Nevertheless, while we made sure that our collective choices strived for a global optimum, our proposed benchmark is not without flaws.

We focus on interpolation with two input frames, which limits the benchmark to a reasonable scope but ignores large areas of the frame interpolation community. At the same time, this limitation follows the paradigm in other areas. Consider optical flow benchmarks for example, they also focus on two-frame inputs even though there is research on multi-frame optical flow estimation. Nevertheless, we hope that others will follow our footsteps and provide dedicated benchmarks for the remaining areas, such as non-linear interpolation. This also includes an analysis of camera nuances such as motion blur and rolling shutter.

Additionally, since we analyze the quality with respect to various per-pixel attributes such as the motion magnitude, we are unfortunately unable to incorporate patch-wise metrics due to their inability to handle sparse inputs.

Lastly, the measured interpolation quality can be boosted through ensembling~\cite{Niklaus:2021:RAC} and we have found the effects of this to be rather severe as shown in \Cref{tbl:ensembling}. For example, SoftSplat ranks fourth without ensembling but it would have ranked first with ensembling when compared to the others without ensembling. We hence ask participants to refrain from using ensembling for a fair comparison while keeping the computational budget low for everyone. Note that we still allow ensembled results as long as it is disclosed.

\begin{figure}\centering
    \setlength{\tabcolsep}{0.0cm}
    \renewcommand{\arraystretch}{1.2}
    \newcommand{\quantTit}[1]{\multicolumn{2}{c}{\scriptsize #1}}
    \newcommand{\quantSec}[1]{\scriptsize #1}
    \newcommand{\quantInd}[1]{\scriptsize #1}
    \newcommand{\quantVal}[1]{\scalebox{0.83}[1.0]{$ #1 $}}
    \newcommand{\quantFirst}[1]{\usolid{\scalebox{0.83}[1.0]{$ #1 $}}}
    \newcommand{\quantSecond}[1]{\udotted{\scalebox{0.83}[1.0]{$ #1 $}}}
    \footnotesize
    \begin{tabularx}{\columnwidth}{@{\hspace{0.1cm}} X P{1.25cm} @{\hspace{-0.4cm}} P{1.12cm} P{1.25cm} @{\hspace{-0.4cm}} P{1.12cm} P{1.25cm} @{\hspace{-0.4cm}} P{1.12cm}}
        \toprule
            & \quantTit{no ensemble} & \quantTit{8 $\times$ ensemble} & \quantTit{16 $\times$ ensemble}
        \\ \cmidrule(l{2pt}r{2pt}){2-3} \cmidrule(l{2pt}r{2pt}){4-5} \cmidrule(l{2pt}r{2pt}){6-7}
            & \quantSec{rank} & \quantSec{$\text{PSNR}^{\ast}$} & \quantSec{rank} & \quantSec{$\text{PSNR}^{\ast}$} & \quantSec{rank} & \quantSec{$\text{PSNR}^{\ast}$}
        \\ \midrule
ABME~\cite{Park:2021:ABM} & \quantVal{\hphantom{0}2\text{ / } 10} & \quantVal{28.63} & \quantVal{\hphantom{0}2\text{ / } 10} & \quantVal{29.38} & \quantVal{\hphantom{0}5\text{ / } 10} & \quantVal{29.46}
\\
AMT-G~\cite{Li:2023:AMT} & \quantVal{\hphantom{0}7\text{ / } 10} & \quantVal{28.18} & \quantVal{\hphantom{0}6\text{ / } 10} & \quantVal{29.04} & \quantVal{\hphantom{0}3\text{ / } 10} & \quantVal{29.50}
\\
CtxSyn~\cite{Niklaus:2018:CAS} & \quantVal{\hphantom{0}1\text{ / } 10} & \quantFirst{29.36} & \quantVal{\hphantom{0}1\text{ / } 10} & \quantFirst{30.29} & \quantVal{\hphantom{0}1\text{ / } 10} & \quantFirst{30.44}
\\
DAIN~\cite{Bao:2019:DAV} & \quantVal{\hphantom{0}8\text{ / } 10} & \quantVal{27.83} & \quantVal{\hphantom{0}8\text{ / } 10} & \quantVal{28.77} & \quantVal{\hphantom{0}7\text{ / } 10} & \quantVal{29.09}
\\
FLDR~\cite{Nottebaum:2022:EFE} & \quantVal{\hphantom{0}9\text{ / } 10} & \quantVal{24.09} & \quantVal{\hphantom{0}9\text{ / } 10} & \quantVal{24.81} & \quantVal{\hphantom{0}9\text{ / } 10} & \quantVal{24.87}
\\
M2M~\cite{Hu:2022:MMS} & \quantVal{\hphantom{0}3\text{ / } 10} & \quantVal{28.61} & \quantVal{\hphantom{0}3\text{ / } 10} & \quantVal{29.32} & \quantVal{\hphantom{0}4\text{ / } 10} & \quantVal{29.48}
\\
SoftSplat~\cite{Niklaus:2020:SSV} & \quantVal{\hphantom{0}4\text{ / } 10} & \quantVal{28.55} & \quantVal{\hphantom{0}3\text{ / } 10} & \quantVal{29.32} & \quantVal{\hphantom{0}2\text{ / } 10} & \quantVal{29.61}
\\
SplatSyn~\cite{Niklaus:2023:SSV} & \quantVal{\hphantom{0}5\text{ / } 10} & \quantVal{28.45} & \quantVal{\hphantom{0}5\text{ / } 10} & \quantVal{29.09} & \quantVal{\hphantom{0}6\text{ / } 10} & \quantVal{29.23}
\\
UPR-Net-L~\cite{Jin:2023:UPR} & \quantVal{\hphantom{0}6\text{ / } 10} & \quantVal{28.31} & \quantVal{\hphantom{0}7\text{ / } 10} & \quantVal{28.84} & \quantVal{\hphantom{0}8\text{ / } 10} & \quantVal{28.95}
\\
XVFI~\cite{Sim:2021:XVF} & \quantVal{10\text{ / } 10} & \quantVal{23.73} & \quantVal{10\text{ / } 10} & \quantVal{24.65} & \quantVal{10\text{ / } 10} & \quantVal{24.78}
        \\ \bottomrule
    \end{tabularx}\vspace{-0.2cm}
    \captionof{table}{Analysis of the impact of performing ensembling~\cite{Niklaus:2021:RAC} at a 1K resolution. We ask participants to refrain from using it.}\vspace{-0.3cm}
    \label{tbl:ensembling}
\end{figure}

\section{Conclusion}
\label{sec:conclusion}

We present a benchmark for evaluating frame interpolation methods. It consists of a carefully designed test dataset with a corresponding evaluation pipeline, which also includes a submission page that ensures consistent results while also facilitating discoverability. Furthermore, thanks to the synthetic nature of our dataset, we have access to rich ground truth information which enables new insights such as the behavior of the interpolation quality with respect to the motion magnitude. Our synthetic data also allows us to make sure that the ground truth follows the constraint of linear motion, avoiding the pitfall in existing test datasets of requiring an oracle. Lastly, our benchmark provides an evaluation of the computational efficiency in a sound manner.

We believe that this benchmark will greatly benefit the frame interpolation community since it offers new insights, allows us to obtain a better understanding of the interpolation space, and opens up areas of where existing methods could be improved. Benchmarks in other areas have helped to tremendously accelerate the research progress and we hope to do the same for frame interpolation.

{
    \small
    \bibliographystyle{ieeenat_fullname}
    \bibliography{main} % \bibliography{bibtex/custom_length,bibtex/papers,bibtex/local}
}

\clearpage
\input{supplementary}

\end{document}

%% file: supplementary.tex
%!TEX root = main.tex

\renewcommand{\thepage}{\roman{page}}
\setcounter{page}{1}
\renewcommand{\thefigure}{A.\arabic{figure}}
\setcounter{figure}{0}
\renewcommand{\thetable}{A.\arabic{table}} 
\setcounter{table}{0}
\renewcommand{\theequation}{A.\arabic{equation}} 
\setcounter{equation}{0}

\appendix
\maketitlesupplementary

\section{Metrics}
\label{appendix:metrics}

\begin{figure*}\centering
    \setlength{\tabcolsep}{0.05cm}
    \setlength{\itemwidth}{3.38cm}
    \hspace*{-\tabcolsep}\begin{tabular}{P{\itemwidth} P{\itemwidth} P{\itemwidth} P{\itemwidth} P{\itemwidth}}
            \includegraphics[]{motionangle-abme}
        &
            \includegraphics[]{motionangle-fldr}
        &
            \includegraphics[]{motionangle-splatsyn}
        &
            \includegraphics[]{motionangle-xvfi}
        &
            \includegraphics[]{motionangle-xvfiv}
        \\
            \includegraphics[]{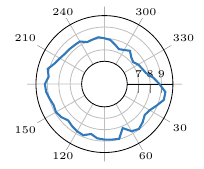}
        &
            \includegraphics[]{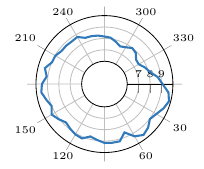}
        &
            \includegraphics[]{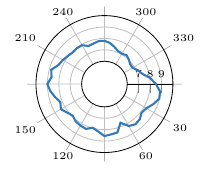}
        &
            \includegraphics[]{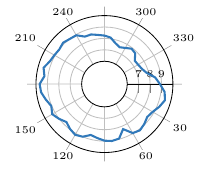}
        &
            \includegraphics[]{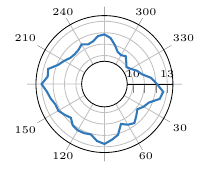}
        \\
            \footnotesize ABME~\cite{Park:2021:ABM}
        &
            \footnotesize FLDR~\cite{Nottebaum:2022:EFE}
        &
            \footnotesize SplatSyn~\cite{Niklaus:2023:SSV}
        &
            \footnotesize XVFI~\cite{Sim:2021:XVF}
        &
            \footnotesize XVFI$_v$~\cite{Sim:2021:XVF}
        \\
    \end{tabular}\vspace{-0.2cm}
    \caption{Analyzing the per-pixel interpolation quality with respect to the motion angle at a 1K resolution using the $\text{PSNR}^{\ast}$ (top) as well as the PSNR (bottom). Each plot indicates the difference between the overall peak signal to noise ratio and the peak signal to noise ratio in each direction. Notice that the plots largely look the same when using the PSNR, which prevents any meaningful insights.}\vspace{-0.0cm}
    \label{fig:metricComparison}
\end{figure*}

To clarify the differences between PSNR and $\text{PSNR}^{\ast}$, our implementations of both metrics are shown in \Cref{fig:psnr,fig:psnrx}. As we already discussed in \Cref{sec:metrics}, we use $\text{PSNR}^{\ast}$ as our metric for measuring the quality of predictions. Due to averaging all individual errors, $\text{PSNR}^{\ast}$ weights each evaluated pixel equally. The effect of using $\text{PSNR}^{\ast}$ instead of PSNR for our evaluation can be seen in \Cref{fig:metricComparison}. It shows how the PSNR values do not contain much information, as the equal weighting of each frame in the PSNR calculation, ignoring the varying number of pixels evaluated per frame for a specific motion angle, leads to a smoothed-out metric. In contrast, our $\text{PSNR}^{\ast}$ clearly shows the performance at each angle and thereby reveals the bias towards horizontal and vertical motions. 

To highlight the similarity between our $\text{PSNR}^{\ast}$ metric and PSNR, we can rewrite \Cref{eq:psnr} as:
\begin{equation}
    \label{eq:psnrRewrite}
    \text{PSNR} = -10 \cdot \log_{10}{\left(\sqrt[N]{\prod \text{MSE}_i}\right)}\ ,
\end{equation}
and assuming that we evaluate $N$ equally sized images, we can rewrite \Cref{eq:psnrx} as:
\begin{equation}
    \label{eq:psnrxRewrite}
    \text{PSNR}^{\ast} = -10 \cdot \log_{10} \left( \frac{1}{N} \sum \text{MSE}_i \right)\ .
\end{equation}
When comparing \Cref{eq:psnrRewrite} and \Cref{eq:psnrxRewrite}, we can see that the only difference is that PSNR calculates the geometric mean over the errors, while $\text{PSNR}^{\ast}$ calculates the arithmetic mean. Due to the inequality of arithmetic and geometric means, PSNR returns bigger or equal values than $\text{PSNR}^{\ast}$. For a more detailed discussion and analysis we refer to Keles et al.~\cite{Keles:2021:CPS}.

\section{Complete Results}

For completeness, we show the single- and multi-frame interpolation results of all of our 21 evaluated methods across multiple resolutions in \Cref{tbl:suppl-main} and \Cref{tbl:suppl-mulframe}. Overall, the findings in these evaluations are consistent with our findings in \Cref{sec:results}. The detailed evaluation results of the interpolation quality with respect to motion magnitudes, angles, and photometric inconsistencies for each of the evaluated 21 methods can be found in \Cref{fig:perpixel-abme,fig:perpixel-adacof,fig:perpixel-amt,fig:perpixel-biformer,fig:perpixel-bmbc,fig:perpixel-cain,fig:perpixel-ctxsyn,fig:perpixel-dain,fig:perpixel-edsc,fig:perpixel-ema,fig:perpixel-fldr,fig:perpixel-m2m,fig:perpixel-rrin,fig:perpixel-sepconv,fig:perpixel-sepconvpp,fig:perpixel-softsplat,fig:perpixel-splatsyn,fig:perpixel-uprnet,fig:perpixel-vfiformer,fig:perpixel-vos,fig:perpixel-xvfi}.

To ensure consistency amongst the runtime measurements in \Cref{tbl:timetable,tbl:suppl-timetable}, we evaluate all methods on a single NVIDIA A100. Details on the software and hardware used in our evaluation system are listed in \Cref{fig:environment}.

\section{Anticipated Questions}
\label{appendix:faq}

\paragraph{Are there moving, non-rigid objects in the benchmark?} No, we cannot produce non-rigid motions due to our construction of the benchmark using static sprites.

\inparagraph{What is the resolution of 1K, 2K, and 4K frames in this benchmark?} 1K, 2K, and 4K correspond to 1024$\times$512, 2048$\times$1024, and 4096$\times$2048 pixels respectively.

\inparagraph{Is the 2-occ. case relevant for real videos?} Yes, using our estimated optical flow on Vimeo-90k, we identified roughly $2\%$ of the interpolated pixels to be occluded in both input frames and $7\%$ of pixels to be occluded in one frame.

\inparagraph{Shouldn't there be two separate cases for 1-occ, because it might make a difference whether an occlusion occurs in the first or second input frame?} We observed no difference between these two cases for any method, so we merged them to lower the number of evaluated cases.

\inparagraph{How do you measure non-linearity?} We utilize the estimated optical flow from the center frame to the first input frame and from the center frame to the second input frame. If a motion is linear, these two flows cancel out when summing them up, as they should point to opposing directions with equal magnitudes. Therefore, we use the 2-norm of this sum to measure the non-linearity.

\inparagraph{Wouldn't rendering realistic images in 3D environments be better to create more complex motions?} From a mathematical standpoint, our approach based on sprites subject to 2D transforms already models all possible types of linear motion and utilizing 3D does not change that (the projection of the 3D motion onto a 2D image plane would yield the same motion types). Furthermore, curating 3D environments that look photo-realistic (take FlyingThings3D as a popular counterexample) is significantly more difficult and laborious than re-purposing and re-mixing real world imagery which is already at least piece-wise photo-realistic.

\inparagraph{Why is there no exploration of more than two inputs?} No benchmark can cover every scenario and we clearly acknowledge this. Take typical optical flow benchmarks for example, they also focus on two-frame inputs even though there is research on multi-frame optical flow estimation.

{
    \small

}

\begin{figure*}\begin{minipage}[b]{.477\textwidth}
\begin{lstlisting}[basicstyle=\scriptsize]
class PSNR(torchmetrics.Metric):
  def __init__(self):
    super().__init__()
    self.add_state('value', default=torch.tensor(0))
    self.add_state('count', default=torch.tensor(0))

  def update(self, pred, target, mask=None):
    target = target.float() / 255
    pred = pred.float() / 255

    se = (target - pred) ** 2
    if mask is None:
      mask = torch.ones_like(se).bool()

    se[~mask] = 0
    mse = se.sum((1, 2, 3)) / mask.sum((1, 2, 3))
    psnr = -10 * torch.log10(mse)
    self.value += psnr.sum()
    self.count += psnr.numel()

  def compute(self):
    return self.value / self.count
\end{lstlisting}\vspace{-0.4cm}
    \caption{Example implementation of PSNR.}
    \label{fig:psnr}
\end{minipage}\qquad\begin{minipage}[b]{.477\textwidth}
\begin{lstlisting}[basicstyle=\scriptsize]
class PSNRX(torchmetrics.Metric):
  def __init__(self):
    super().__init__()
    self.add_state('value', default=torch.tensor(0))
    self.add_state('count', default=torch.tensor(0))

  def update(self, pred, target, mask=None):
    target = target.float() / 255
    pred = pred.float() / 255

    se = (target - pred) ** 2
    if mask is not None:
      se = se[mask]

    self.value += se.sum()
    self.count += se.numel()

  def compute(self):
    mse = self.value / self.count
    return -10 * torch.log10(mse)
\end{lstlisting}\vspace{-0.4cm}
    \caption{Example implementation of $\text{PSNR}^{\ast}$.}
    \label{fig:psnrx}
\end{minipage}\end{figure*}

\begin{figure*}\centering
    \setlength{\tabcolsep}{0.05cm}
    \setlength{\itemwidth}{3.4cm} % 1.855cm
    \hspace*{-\tabcolsep}\begin{tabular}{ccccc}
            \includegraphics[width=\itemwidth]{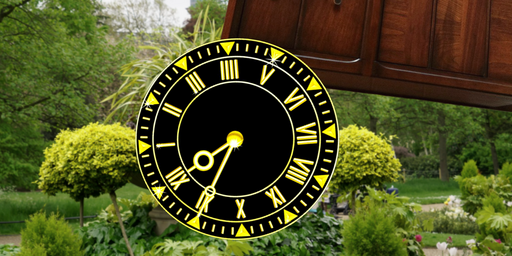}
        &
            \includegraphics[width=\itemwidth]{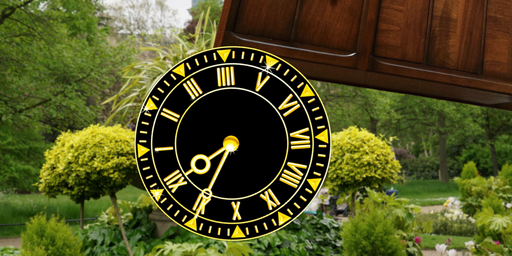}
        &
            \includegraphics[width=\itemwidth]{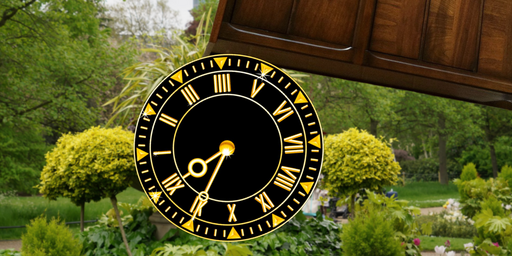}
        &
            \includegraphics[width=\itemwidth]{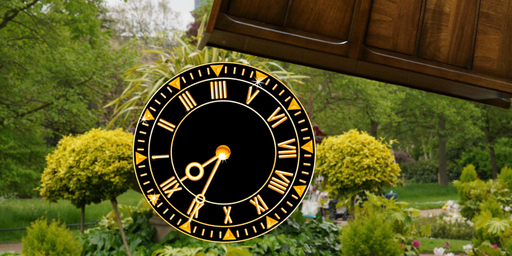}
        &
            \includegraphics[width=\itemwidth]{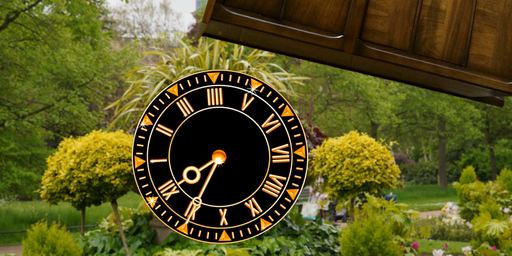}
        \\
            \includegraphics[width=\itemwidth]{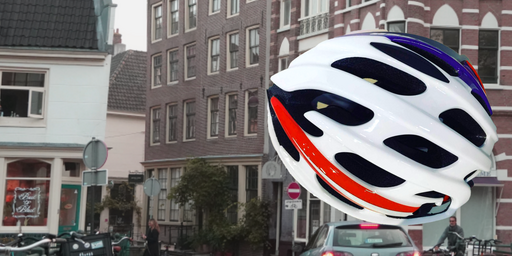}
        &
            \includegraphics[width=\itemwidth]{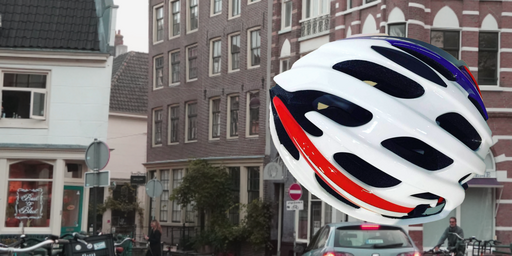}
        &
            \includegraphics[width=\itemwidth]{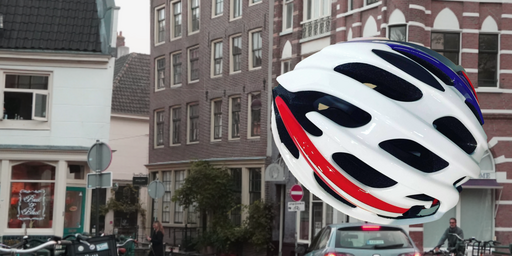}
        &
            \includegraphics[width=\itemwidth]{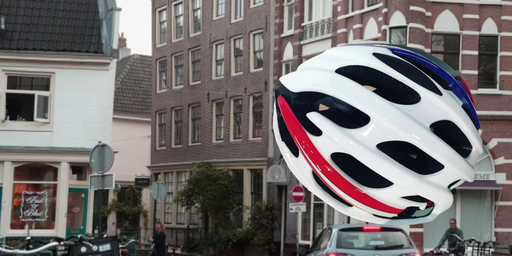}
        &
            \includegraphics[width=\itemwidth]{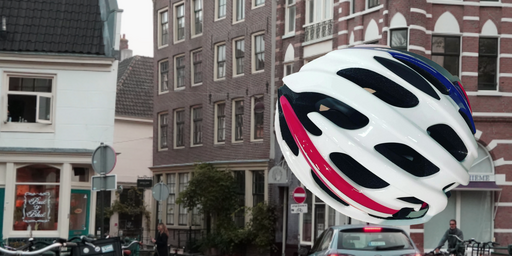}
        \\
            \includegraphics[width=\itemwidth]{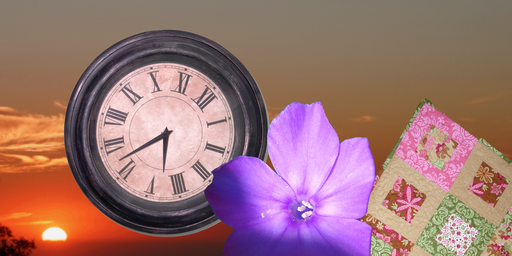}
        &
            \includegraphics[width=\itemwidth]{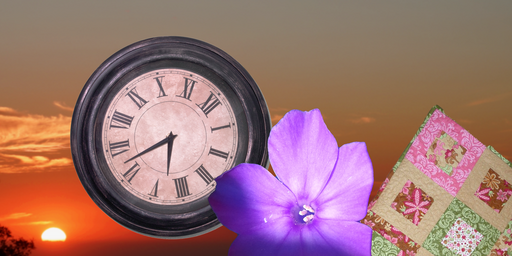}
        &
            \includegraphics[width=\itemwidth]{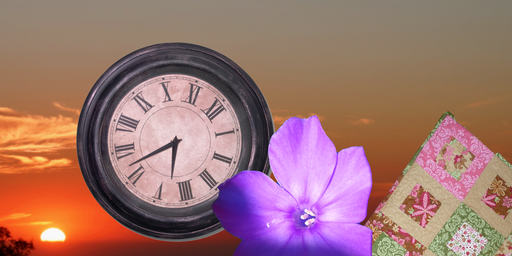}
        &
            \includegraphics[width=\itemwidth]{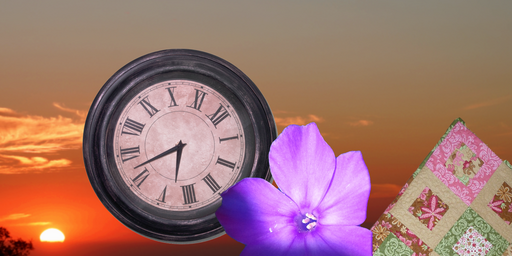}
        &
            \includegraphics[width=\itemwidth]{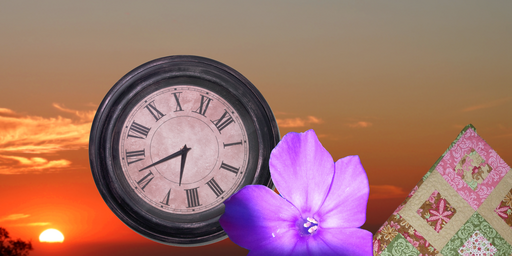}
        \\
            \includegraphics[width=\itemwidth]{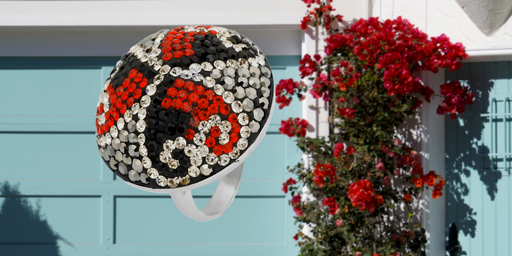}
        &
            \includegraphics[width=\itemwidth]{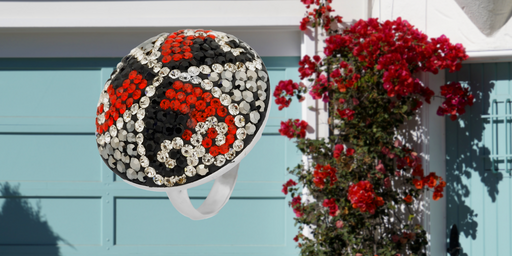}
        &
            \includegraphics[width=\itemwidth]{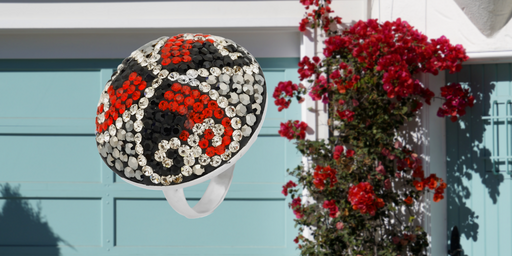}
        &
            \includegraphics[width=\itemwidth]{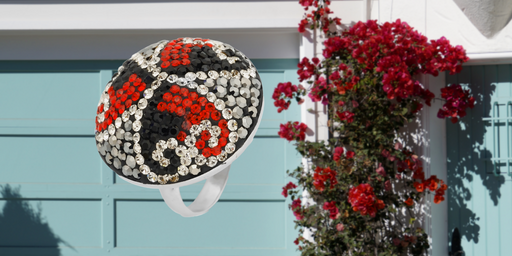}
        &
            \includegraphics[width=\itemwidth]{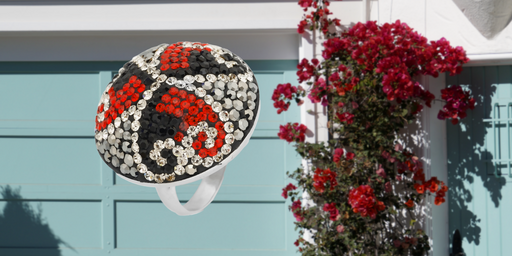}
        \\
            \includegraphics[width=\itemwidth]{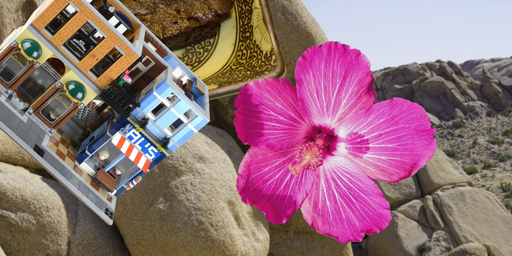}
        &
            \includegraphics[width=\itemwidth]{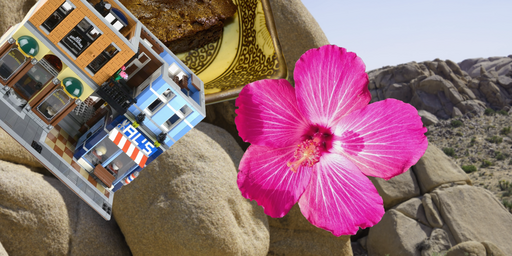}
        &
            \includegraphics[width=\itemwidth]{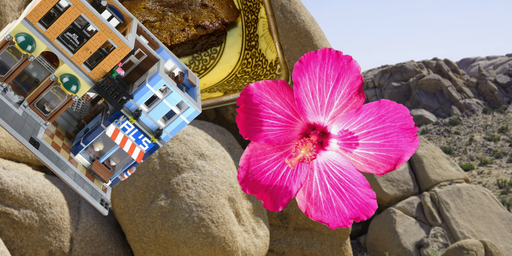}
        &
            \includegraphics[width=\itemwidth]{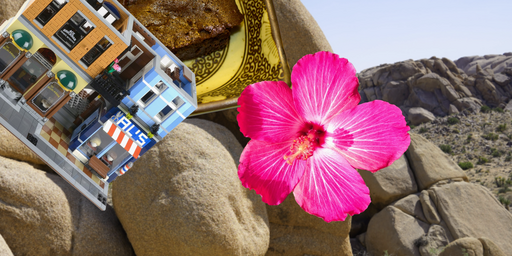}
        &
            \includegraphics[width=\itemwidth]{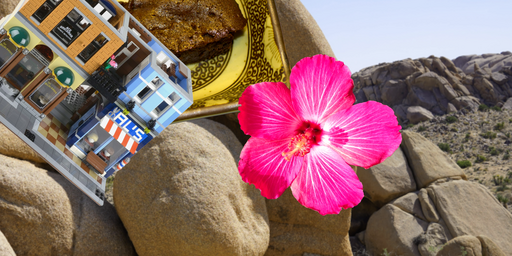}
        \\
            \footnotesize $t = 0.0$
        &
            \footnotesize $t = 0.25$
        &
            \footnotesize $t = 0.5$
        &
            \footnotesize $t = 0.75$
        &
            \footnotesize $t = 1.0$
        \\
    \end{tabular}\vspace{-0.2cm}
    \caption{Samples from our benchmarking datasets, participants are given $t = 0$ and $t = 1$ while the in-between frames are withheld.}\vspace{-0.0cm}
    \label{fig:samples}
\end{figure*}

\begin{figure*}\centering
    \setlength{\tabcolsep}{0.0cm}
    \renewcommand{\arraystretch}{1.2}
    \newcommand{\quantTit}[1]{\multicolumn{5}{c}{\scriptsize #1}}
    \newcommand{\quantSec}[1]{\scriptsize #1}
    \newcommand{\quantVal}[1]{\scalebox{0.83}[1.0]{$ #1 $}}
    \newcommand{\quantFirst}[1]{\usolid{\scalebox{0.83}[1.0]{$ #1 $}}}
    \newcommand{\quantSecond}[1]{\udotted{\scalebox{0.83}[1.0]{$ #1 $}}}
    \footnotesize
    \begin{tabularx}{\textwidth}{@{\hspace{0.1cm}} X P{1.3cm} @{\hspace{-0.3cm}} P{1.2cm} @{\hspace{-0.3cm}} P{1.2cm} @{\hspace{-0.3cm}} P{1.2cm} @{\hspace{-0.3cm}} P{1.2cm} P{1.3cm} @{\hspace{-0.3cm}} P{1.2cm} @{\hspace{-0.3cm}} P{1.2cm} @{\hspace{-0.3cm}} P{1.2cm} @{\hspace{-0.3cm}} P{1.2cm} P{1.3cm} @{\hspace{-0.3cm}} P{1.2cm} @{\hspace{-0.3cm}} P{1.2cm} @{\hspace{-0.3cm}} P{1.2cm} @{\hspace{-0.3cm}} P{1.2cm}}
        \toprule
             & \quantTit{1K} & \quantTit{2K} & \quantTit{4K}
        \\ \cmidrule(l{2pt}r{2pt}){2-6} \cmidrule(l{2pt}r{2pt}){7-11} \cmidrule(l{2pt}r{2pt}){12-16}
             & \quantSec{rank} & \quantSec{all} & \quantSec{0-occ.} & \quantSec{1-occ.} & \quantSec{2-occ.} & \quantSec{rank} & \quantSec{all} & \quantSec{0-occ.} & \quantSec{1-occ.} & \quantSec{2-occ.} & \quantSec{rank} & \quantSec{all} & \quantSec{0-occ.} & \quantSec{1-occ.} & \quantSec{2-occ.}
        \\ \midrule
ABME~\cite{Park:2021:ABM} & \quantVal{\hphantom{0}2\text{ / } 21} & \quantVal{28.63} & \quantVal{31.62} & \quantVal{21.36} & \quantVal{16.95} & \quantVal{\hphantom{0}5\text{ / } 21} & \quantVal{23.95} & \quantVal{25.35} & \quantVal{18.41} & \quantVal{15.54} & \quantVal{10\text{ / } 21} & \quantVal{18.35} & \quantVal{18.80} & \quantVal{15.41} & \quantVal{13.97}
\\
AdaCoF~\cite{Lee:2020:ACF} & \quantVal{20\text{ / } 21} & \quantVal{20.44} & \quantVal{21.10} & \quantVal{16.83} & \quantVal{15.40} & \quantVal{20\text{ / } 21} & \quantVal{18.27} & \quantVal{18.67} & \quantVal{15.62} & \quantVal{14.42} & \quantVal{16\text{ / } 21} & \quantVal{17.09} & \quantVal{17.35} & \quantVal{15.13} & \quantVal{13.87}
\\
AMT-G~\cite{Li:2023:AMT} & \quantVal{\hphantom{0}7\text{ / } 21} & \quantVal{28.18} & \quantVal{30.62} & \quantVal{21.35} & \quantFirst{17.80} & \quantVal{11\text{ / } 21} & \quantVal{20.97} & \quantVal{21.71} & \quantVal{17.00} & \quantVal{15.72} & \quantVal{-} & \quantVal{-} & \quantVal{-} & \quantVal{-} & \quantVal{-}
\\
BiFormer~\cite{Park:2023:LBM} & \quantVal{-} & \quantVal{-} & \quantVal{-} & \quantVal{-} & \quantVal{-} & \quantVal{-} & \quantVal{-} & \quantVal{-} & \quantVal{-} & \quantVal{-} & \quantVal{\hphantom{0}2\text{ / } 21} & \quantVal{26.78} & \quantVal{29.55} & \quantVal{19.35} & \quantVal{15.55}
\\
BMBC~\cite{Park:2020:BME} & \quantVal{14\text{ / } 21} & \quantVal{22.18} & \quantVal{23.14} & \quantVal{17.74} & \quantVal{15.16} & \quantVal{19\text{ / } 21} & \quantVal{18.53} & \quantVal{19.04} & \quantVal{15.43} & \quantVal{14.26} & \quantVal{-} & \quantVal{-} & \quantVal{-} & \quantVal{-} & \quantVal{-}
\\
CAIN~\cite{Choi:2020:CAA} & \quantVal{18\text{ / } 21} & \quantVal{21.83} & \quantVal{22.69} & \quantVal{17.64} & \quantVal{15.92} & \quantVal{17\text{ / } 21} & \quantVal{18.95} & \quantVal{19.47} & \quantVal{15.80} & \quantVal{14.79} & \quantVal{13\text{ / } 21} & \quantVal{17.67} & \quantVal{18.04} & \quantVal{15.09} & \quantVal{14.03}
\\
CtxSyn~\cite{Niklaus:2018:CAS} & \quantVal{\hphantom{0}1\text{ / } 21} & \quantFirst{29.36} & \quantFirst{32.72} & \quantVal{21.81} & \quantVal{17.34} & \quantVal{\hphantom{0}7\text{ / } 21} & \quantVal{23.33} & \quantVal{24.65} & \quantVal{17.91} & \quantVal{15.81} & \quantVal{\hphantom{0}9\text{ / } 21} & \quantVal{18.80} & \quantVal{19.47} & \quantVal{14.98} & \quantVal{13.75}
\\
DAIN~\cite{Bao:2019:DAV} & \quantVal{\hphantom{0}8\text{ / } 21} & \quantVal{27.83} & \quantVal{31.47} & \quantVal{19.97} & \quantVal{16.94} & \quantVal{\hphantom{0}3\text{ / } 21} & \quantVal{26.81} & \quantVal{29.77} & \quantVal{19.32} & \quantVal{16.40} & \quantVal{-} & \quantVal{-} & \quantVal{-} & \quantVal{-} & \quantVal{-}
\\
EDSC~\cite{Cheng:2022:MVF} & \quantVal{13\text{ / } 21} & \quantVal{22.90} & \quantVal{23.91} & \quantVal{18.29} & \quantVal{16.35} & \quantVal{11\text{ / } 21} & \quantVal{20.97} & \quantVal{21.60} & \quantVal{17.38} & \quantVal{15.91} & \quantVal{\hphantom{0}7\text{ / } 21} & \quantVal{19.14} & \quantVal{19.57} & \quantVal{16.32} & \quantVal{15.02}
\\
EMA~\cite{Zhang:2023:EMA} & \quantVal{\hphantom{0}9\text{ / } 21} & \quantVal{25.95} & \quantVal{27.36} & \quantVal{20.53} & \quantVal{17.68} & \quantVal{10\text{ / } 21} & \quantVal{21.29} & \quantVal{21.94} & \quantVal{17.61} & \quantVal{16.17} & \quantVal{-} & \quantVal{-} & \quantVal{-} & \quantVal{-} & \quantVal{-}
\\
FLDR~\cite{Nottebaum:2022:EFE} & \quantVal{10\text{ / } 21} & \quantVal{24.09} & \quantVal{25.80} & \quantVal{18.14} & \quantVal{15.56} & \quantVal{\hphantom{0}6\text{ / } 21} & \quantVal{23.80} & \quantVal{25.38} & \quantVal{17.96} & \quantVal{15.12} & \quantVal{\hphantom{0}4\text{ / } 21} & \quantVal{23.53} & \quantVal{25.05} & \quantVal{17.68} & \quantVal{14.47}
\\
M2M~\cite{Hu:2022:MMS} & \quantVal{\hphantom{0}3\text{ / } 21} & \quantVal{28.61} & \quantVal{31.73} & \quantVal{21.21} & \quantVal{17.16} & \quantVal{\hphantom{0}4\text{ / } 21} & \quantVal{25.81} & \quantVal{28.12} & \quantVal{18.95} & \quantVal{15.99} & \quantVal{\hphantom{0}6\text{ / } 21} & \quantVal{20.61} & \quantVal{21.44} & \quantVal{16.29} & \quantVal{14.49}
\\
RRIN~\cite{Li:2020:VFI} & \quantVal{15\text{ / } 21} & \quantVal{22.17} & \quantVal{23.08} & \quantVal{17.81} & \quantVal{16.09} & \quantVal{14\text{ / } 21} & \quantVal{19.39} & \quantVal{19.94} & \quantVal{16.10} & \quantVal{15.17} & \quantVal{14\text{ / } 21} & \quantVal{17.61} & \quantVal{18.04} & \quantVal{14.73} & \quantVal{13.75}
\\
SepConv~\cite{Niklaus:2017:VFIAS} & \quantVal{16\text{ / } 21} & \quantVal{22.00} & \quantVal{23.06} & \quantVal{17.25} & \quantVal{15.86} & \quantVal{15\text{ / } 21} & \quantVal{19.35} & \quantVal{19.90} & \quantVal{16.05} & \quantVal{15.17} & \quantVal{12\text{ / } 21} & \quantVal{17.79} & \quantVal{18.16} & \quantVal{15.24} & \quantVal{14.28}
\\
SepConv++~\cite{Niklaus:2021:RAC} & \quantVal{17\text{ / } 21} & \quantVal{21.85} & \quantVal{22.64} & \quantVal{17.84} & \quantVal{16.28} & \quantVal{16\text{ / } 21} & \quantVal{19.23} & \quantVal{19.74} & \quantVal{16.08} & \quantVal{15.25} & \quantVal{11\text{ / } 21} & \quantVal{17.84} & \quantVal{18.24} & \quantVal{15.11} & \quantVal{14.07}
\\
SoftSplat~\cite{Niklaus:2020:SSV} & \quantVal{\hphantom{0}4\text{ / } 21} & \quantVal{28.55} & \quantVal{31.28} & \quantVal{21.46} & \quantVal{17.65} & \quantVal{\hphantom{0}9\text{ / } 21} & \quantVal{22.59} & \quantVal{23.58} & \quantVal{17.91} & \quantVal{16.05} & \quantVal{\hphantom{0}8\text{ / } 21} & \quantVal{18.85} & \quantVal{19.45} & \quantVal{15.28} & \quantVal{14.04}
\\
SplatSyn~\cite{Niklaus:2023:SSV} & \quantVal{\hphantom{0}5\text{ / } 21} & \quantVal{28.45} & \quantVal{30.89} & \quantVal{21.71} & \quantVal{17.20} & \quantVal{\hphantom{0}1\text{ / } 21} & \quantFirst{28.40} & \quantFirst{30.91} & \quantFirst{21.43} & \quantFirst{16.71} & \quantVal{\hphantom{0}1\text{ / } 21} & \quantFirst{27.40} & \quantFirst{29.71} & \quantFirst{20.48} & \quantFirst{15.97}
\\
UPR-Net-L~\cite{Jin:2023:UPR} & \quantVal{\hphantom{0}6\text{ / } 21} & \quantVal{28.31} & \quantVal{30.28} & \quantFirst{22.14} & \quantVal{17.46} & \quantVal{\hphantom{0}2\text{ / } 21} & \quantVal{27.09} & \quantVal{28.76} & \quantVal{21.20} & \quantVal{16.47} & \quantVal{\hphantom{0}3\text{ / } 21} & \quantVal{25.69} & \quantVal{27.05} & \quantVal{20.28} & \quantVal{14.65}
\\
VFIformer~\cite{Lu:2022:VFI} & \quantVal{11\text{ / } 21} & \quantVal{23.83} & \quantVal{24.79} & \quantVal{19.35} & \quantVal{17.26} & \quantVal{13\text{ / } 21} & \quantVal{19.91} & \quantVal{20.48} & \quantVal{16.56} & \quantVal{15.68} & \quantVal{-} & \quantVal{-} & \quantVal{-} & \quantVal{-} & \quantVal{-}
\\
VOS~\cite{Yoo:2023:VOS} & \quantVal{19\text{ / } 21} & \quantVal{20.95} & \quantVal{21.70} & \quantVal{17.06} & \quantVal{15.74} & \quantVal{18\text{ / } 21} & \quantVal{18.74} & \quantVal{19.29} & \quantVal{15.44} & \quantVal{14.73} & \quantVal{15\text{ / } 21} & \quantVal{17.38} & \quantVal{17.90} & \quantVal{14.11} & \quantVal{13.38}
\\
XVFI~\cite{Sim:2021:XVF} & \quantVal{12\text{ / } 21} & \quantVal{23.73} & \quantVal{25.36} & \quantVal{17.90} & \quantVal{15.33} & \quantVal{\hphantom{0}8\text{ / } 21} & \quantVal{23.25} & \quantVal{24.74} & \quantVal{17.54} & \quantVal{14.95} & \quantVal{\hphantom{0}5\text{ / } 21} & \quantVal{22.48} & \quantVal{23.71} & \quantVal{17.15} & \quantVal{14.33}
        \\ \bottomrule
    \end{tabularx}\vspace{-0.2cm}
    \captionof{table}{Results of all evaluated frame interpolation methods across multiple resolutions. We report the $\text{PSNR}^{\ast}$ separately across all areas (all), non-occluded areas (0-occ.), areas that are occluded in one input (1-occ.), and areas that are occluded in both inputs (2-occ.). We are unfortunately unable to report some numbers denoted with a dash due to not having enough GPU memory to run these methods or due to minimum resolution requirements of the methods.}\vspace{-0.0cm}
    \label{tbl:suppl-main}
\end{figure*}

\begin{figure*}\centering
    \setlength{\tabcolsep}{0.0cm}
    \renewcommand{\arraystretch}{1.2}
    \newcommand{\quantTit}[1]{\multicolumn{8}{c}{\scriptsize #1}}
    \newcommand{\quantSec}[1]{\scriptsize \scalebox{0.78}[1.0]{$ #1 $}}
    \newcommand{\quantVal}[1]{\scalebox{0.83}[1.0]{$ #1 $}}
    \newcommand{\quantFirst}[1]{\usolid{\scalebox{0.83}[1.0]{$ #1 $}}}
    \newcommand{\quantSecond}[1]{\udotted{\scalebox{0.83}[1.0]{$ #1 $}}}
    \footnotesize
    \begin{tabularx}{\textwidth}{@{\hspace{0.1cm}} X P{1.3cm} @{\hspace{-0.3cm}} P{1.17cm} @{\hspace{-0.3cm}} P{1.17cm} @{\hspace{-0.3cm}} P{1.17cm} @{\hspace{-0.3cm}} P{1.17cm} @{\hspace{-0.3cm}} P{1.17cm} @{\hspace{-0.3cm}} P{1.17cm} @{\hspace{-0.3cm}} P{1.17cm} P{1.3cm} @{\hspace{-0.3cm}} P{1.17cm} @{\hspace{-0.3cm}} P{1.17cm} @{\hspace{-0.3cm}} P{1.17cm} @{\hspace{-0.3cm}} P{1.17cm} @{\hspace{-0.3cm}} P{1.17cm} @{\hspace{-0.3cm}} P{1.17cm} @{\hspace{-0.3cm}} P{1.17cm}}
        \toprule
             & \quantTit{$\text{PSNR}^{\ast}$} & \quantTit{$\text{PSNR}^{\ast}_{\sigma}$}
        \\ \cmidrule(l{2pt}r{2pt}){2-9} \cmidrule(l{2pt}r{2pt}){10-17} 
             & \scriptsize rank & \quantSec{$t = 0.125$} & \quantSec{$t = 0.25$} & \quantSec{$t = 0.375$} & \quantSec{$t = 0.5$} & \quantSec{$t = 0.625$} & \quantSec{$t = 0.75$} & \quantSec{$t = 0.875$} & \scriptsize rank & \quantSec{$t = 0.125$} & \quantSec{$t = 0.25$} & \quantSec{$t = 0.375$} & \quantSec{$t = 0.5$} & \quantSec{$t = 0.625$} & \quantSec{$t = 0.75$} & \quantSec{$t = 0.875$}
        \\ \midrule
ABME~\cite{Park:2021:ABM} & \quantVal{\hphantom{0}3\text{ / } 20} & \quantVal{31.00} & \quantVal{29.43} & \quantVal{28.99} & \quantVal{28.63} & \quantVal{28.67} & \quantVal{28.91} & \quantVal{30.57} & \quantVal{\hphantom{0}3\text{ / } 20} & \quantVal{21.38} & \quantVal{20.26} & \quantVal{20.05} & \quantVal{19.72} & \quantVal{19.86} & \quantVal{19.95} & \quantVal{21.02}
\\
AdaCoF~\cite{Lee:2020:ACF} & \quantVal{18\text{ / } 20} & \quantVal{23.65} & \quantVal{21.83} & \quantVal{21.10} & \quantVal{20.44} & \quantVal{20.98} & \quantVal{21.58} & \quantVal{23.11} & \quantVal{18\text{ / } 20} & \quantVal{17.00} & \quantVal{15.79} & \quantVal{15.33} & \quantVal{14.83} & \quantVal{15.23} & \quantVal{15.59} & \quantVal{16.58}
\\
AMT-G~\cite{Li:2023:AMT} & \quantVal{20\text{ / } 20} & \quantVal{16.69} & \quantVal{17.68} & \quantVal{19.44} & \quantVal{28.18} & \quantVal{19.47} & \quantVal{17.37} & \quantVal{16.29} & \quantVal{20\text{ / } 20} & \quantVal{12.14} & \quantVal{12.74} & \quantVal{13.86} & \quantVal{19.28} & \quantVal{13.88} & \quantVal{12.50} & \quantVal{11.86}
\\
BiFormer~\cite{Park:2023:LBM} & \quantVal{-} & \quantVal{-} & \quantVal{-} & \quantVal{-} & \quantVal{-} & \quantVal{-} & \quantVal{-} & \quantVal{-} & \quantVal{-} & \quantVal{-} & \quantVal{-} & \quantVal{-} & \quantVal{-} & \quantVal{-} & \quantVal{-} & \quantVal{-}
\\
BMBC~\cite{Park:2020:BME} & \quantVal{16\text{ / } 20} & \quantVal{24.31} & \quantVal{23.13} & \quantVal{22.64} & \quantVal{22.18} & \quantVal{22.43} & \quantVal{22.89} & \quantVal{24.09} & \quantVal{17\text{ / } 20} & \quantVal{17.19} & \quantVal{16.13} & \quantVal{15.69} & \quantVal{15.37} & \quantVal{15.52} & \quantVal{15.90} & \quantVal{16.98}
\\
CAIN~\cite{Choi:2020:CAA} & \quantVal{15\text{ / } 20} & \quantVal{25.27} & \quantVal{23.30} & \quantVal{22.44} & \quantVal{21.83} & \quantVal{22.37} & \quantVal{23.19} & \quantVal{25.04} & \quantVal{14\text{ / } 20} & \quantVal{17.88} & \quantVal{16.66} & \quantVal{16.17} & \quantVal{15.84} & \quantVal{16.10} & \quantVal{16.55} & \quantVal{17.68}
\\
CtxSyn~\cite{Niklaus:2018:CAS} & \quantVal{\hphantom{0}5\text{ / } 20} & \quantVal{28.72} & \quantVal{29.07} & \quantFirst{29.30} & \quantFirst{29.36} & \quantFirst{29.26} & \quantVal{29.14} & \quantVal{29.11} & \quantVal{\hphantom{0}1\text{ / } 20} & \quantVal{21.87} & \quantFirst{21.19} & \quantFirst{20.77} & \quantFirst{20.55} & \quantFirst{20.56} & \quantFirst{20.85} & \quantFirst{21.61}
\\
DAIN~\cite{Bao:2019:DAV} & \quantVal{\hphantom{0}7\text{ / } 20} & \quantVal{27.14} & \quantVal{26.81} & \quantVal{27.15} & \quantVal{27.83} & \quantVal{27.05} & \quantVal{26.88} & \quantVal{27.55} & \quantVal{\hphantom{0}8\text{ / } 20} & \quantVal{19.51} & \quantVal{18.68} & \quantVal{18.59} & \quantVal{18.71} & \quantVal{18.55} & \quantVal{18.73} & \quantVal{19.80}
\\
EDSC~\cite{Cheng:2022:MVF} & \quantVal{12\text{ / } 20} & \quantVal{25.71} & \quantVal{23.52} & \quantVal{23.09} & \quantVal{22.90} & \quantVal{23.05} & \quantVal{23.60} & \quantVal{25.33} & \quantVal{12\text{ / } 20} & \quantVal{18.19} & \quantVal{16.85} & \quantVal{16.70} & \quantVal{16.65} & \quantVal{16.68} & \quantVal{16.87} & \quantVal{18.00}
\\
EMA~\cite{Zhang:2023:EMA} & \quantVal{\hphantom{0}8\text{ / } 20} & \quantVal{29.17} & \quantVal{27.26} & \quantVal{26.45} & \quantVal{25.95} & \quantVal{26.29} & \quantVal{26.99} & \quantVal{28.75} & \quantVal{\hphantom{0}7\text{ / } 20} & \quantVal{20.32} & \quantVal{19.11} & \quantVal{18.66} & \quantVal{18.38} & \quantVal{18.51} & \quantVal{18.86} & \quantVal{20.00}
\\
FLDR~\cite{Nottebaum:2022:EFE} & \quantVal{\hphantom{0}9\text{ / } 20} & \quantVal{27.30} & \quantVal{25.24} & \quantVal{24.38} & \quantVal{24.09} & \quantVal{24.26} & \quantVal{24.97} & \quantVal{26.82} & \quantVal{\hphantom{0}9\text{ / } 20} & \quantVal{19.17} & \quantVal{17.84} & \quantVal{17.37} & \quantVal{17.20} & \quantVal{17.27} & \quantVal{17.68} & \quantVal{18.92}
\\
M2M~\cite{Hu:2022:MMS} & \quantVal{\hphantom{0}1\text{ / } 20} & \quantFirst{31.83} & \quantFirst{29.86} & \quantVal{28.93} & \quantVal{28.61} & \quantVal{28.57} & \quantVal{29.19} & \quantVal{30.89} & \quantVal{\hphantom{0}4\text{ / } 20} & \quantVal{21.94} & \quantVal{20.53} & \quantVal{19.87} & \quantVal{19.52} & \quantVal{19.48} & \quantVal{19.81} & \quantVal{20.99}
\\
RRIN~\cite{Li:2020:VFI} & \quantVal{19\text{ / } 20} & \quantVal{21.18} & \quantVal{21.45} & \quantVal{21.92} & \quantVal{22.17} & \quantVal{21.92} & \quantVal{21.49} & \quantVal{21.15} & \quantVal{19\text{ / } 20} & \quantVal{15.03} & \quantVal{15.25} & \quantVal{15.56} & \quantVal{15.65} & \quantVal{15.51} & \quantVal{15.32} & \quantVal{15.11}
\\
SepConv~\cite{Niklaus:2017:VFIAS} & \quantVal{13\text{ / } 20} & \quantVal{25.67} & \quantVal{23.62} & \quantVal{22.68} & \quantVal{22.00} & \quantVal{22.54} & \quantVal{23.34} & \quantVal{25.12} & \quantVal{15\text{ / } 20} & \quantVal{17.84} & \quantVal{16.66} & \quantVal{16.14} & \quantVal{15.72} & \quantVal{16.01} & \quantVal{16.41} & \quantVal{17.38}
\\
SepConv++~\cite{Niklaus:2021:RAC} & \quantVal{14\text{ / } 20} & \quantVal{25.08} & \quantVal{23.43} & \quantVal{22.54} & \quantVal{21.85} & \quantVal{22.47} & \quantVal{23.27} & \quantVal{24.74} & \quantVal{13\text{ / } 20} & \quantVal{18.07} & \quantVal{16.94} & \quantVal{16.36} & \quantVal{15.85} & \quantVal{16.30} & \quantVal{16.82} & \quantVal{17.84}
\\
SoftSplat~\cite{Niklaus:2020:SSV} & \quantVal{\hphantom{0}4\text{ / } 20} & \quantVal{30.78} & \quantVal{29.30} & \quantVal{28.79} & \quantVal{28.55} & \quantVal{28.52} & \quantVal{28.78} & \quantVal{30.04} & \quantVal{\hphantom{0}5\text{ / } 20} & \quantVal{21.19} & \quantVal{20.19} & \quantVal{19.82} & \quantVal{19.62} & \quantVal{19.59} & \quantVal{19.73} & \quantVal{20.53}
\\
SplatSyn~\cite{Niklaus:2023:SSV} & \quantVal{\hphantom{0}2\text{ / } 20} & \quantVal{31.78} & \quantVal{29.72} & \quantVal{28.70} & \quantVal{28.45} & \quantVal{28.53} & \quantFirst{29.30} & \quantFirst{31.02} & \quantVal{\hphantom{0}2\text{ / } 20} & \quantFirst{22.32} & \quantVal{20.87} & \quantVal{20.23} & \quantVal{19.97} & \quantVal{19.96} & \quantVal{20.29} & \quantVal{21.50}
\\
UPR-Net-L~\cite{Jin:2023:UPR} & \quantVal{\hphantom{0}6\text{ / } 20} & \quantVal{27.90} & \quantVal{27.85} & \quantVal{28.34} & \quantVal{28.31} & \quantVal{28.03} & \quantVal{27.69} & \quantVal{27.90} & \quantVal{\hphantom{0}6\text{ / } 20} & \quantVal{19.06} & \quantVal{19.38} & \quantVal{19.91} & \quantVal{19.77} & \quantVal{19.69} & \quantVal{19.46} & \quantVal{19.70}
\\
VFIformer~\cite{Lu:2022:VFI} & \quantVal{10\text{ / } 20} & \quantVal{26.88} & \quantVal{25.07} & \quantVal{24.31} & \quantVal{23.83} & \quantVal{24.10} & \quantVal{24.75} & \quantVal{26.60} & \quantVal{11\text{ / } 20} & \quantVal{18.45} & \quantVal{17.33} & \quantVal{16.92} & \quantVal{16.61} & \quantVal{16.73} & \quantVal{17.08} & \quantVal{18.28}
\\
VOS~\cite{Yoo:2023:VOS} & \quantVal{17\text{ / } 20} & \quantVal{24.38} & \quantVal{22.58} & \quantVal{21.68} & \quantVal{20.95} & \quantVal{21.60} & \quantVal{22.47} & \quantVal{24.36} & \quantVal{16\text{ / } 20} & \quantVal{17.38} & \quantVal{16.19} & \quantVal{15.63} & \quantVal{15.13} & \quantVal{15.57} & \quantVal{16.10} & \quantVal{17.34}
\\
XVFI~\cite{Sim:2021:XVF} & \quantVal{11\text{ / } 20} & \quantVal{26.43} & \quantVal{24.71} & \quantVal{23.99} & \quantVal{23.73} & \quantVal{23.83} & \quantVal{24.39} & \quantVal{25.85} & \quantVal{10\text{ / } 20} & \quantVal{18.54} & \quantVal{17.39} & \quantVal{16.95} & \quantVal{16.79} & \quantVal{16.85} & \quantVal{17.20} & \quantVal{18.16}
        \\ \bottomrule
    \end{tabularx}\vspace{-0.2cm}
    \captionof{table}{Evaluation of the ability to perform multi-frame interpolation based on seven in-between ground truth frames at a 1K resolution.}\vspace{-0.0cm}
    \label{tbl:suppl-mulframe}
\end{figure*}

\begin{figure*}\begin{minipage}[b]{.477\textwidth}
    \setlength{\tabcolsep}{0.0cm}
    \renewcommand{\arraystretch}{1.2}
    \newcommand{\quantTit}[1]{\multicolumn{3}{c}{\scriptsize #1}}
    \newcommand{\quantSec}[1]{\scriptsize #1}
    \newcommand{\quantInd}[1]{\scriptsize #1}
    \newcommand{\quantVal}[1]{\scalebox{0.83}[1.0]{$ #1 $}}
    \newcommand{\quantFirst}[1]{\usolid{\scalebox{0.83}[1.0]{$ #1 $}}}
    \newcommand{\quantSecond}[1]{\udotted{\scalebox{0.83}[1.0]{$ #1 $}}}
    \footnotesize
    \begin{tabularx}{\columnwidth}{@{\hspace{0.1cm}} X P{1.05cm} @{\hspace{-0.1cm}} P{1.05cm} @{\hspace{-0.1cm}} P{1.05cm} P{1.05cm} @{\hspace{-0.1cm}} P{1.05cm} @{\hspace{-0.1cm}} P{1.05cm}}
        \toprule
            & \quantTit{1 frame} & \quantTit{7 frames}
        \\ \cmidrule(l{2pt}r{2pt}){2-4} \cmidrule(l{2pt}r{2pt}){5-7}
            & {\scriptsize 1K} & {\scriptsize 2K} & {\scriptsize 4K} & {\scriptsize 1K} & {\scriptsize 2K} & {\scriptsize 4K}
        \\ \midrule
ABME~\cite{Park:2021:ABM} & \quantVal{0.29} & \quantVal{1.08} & \quantVal{4.34} & \quantVal{\hphantom{0}}\quantVal{2.00} & \quantVal{\hphantom{0}}\quantVal{7.88} & \quantVal{30.51}
\\
AdaCoF~\cite{Lee:2020:ACF} & \quantFirst{0.02} & \quantVal{0.06} & \quantVal{0.23} & \quantVal{\hphantom{0}}\quantVal{0.12} & \quantVal{\hphantom{0}}\quantVal{0.42} & \quantVal{\hphantom{0}}\quantVal{1.62}
\\
AMT-G~\cite{Li:2023:AMT} & \quantVal{0.07} & \quantVal{0.31} & \quantVal{-} & \quantVal{\hphantom{0}}\quantVal{0.48} & \quantVal{\hphantom{0}}\quantVal{2.20} & \quantVal{-}
\\
BiFormer~\cite{Park:2023:LBM} & \quantVal{-} & \quantVal{-} & \quantVal{1.92} & \quantVal{-} & \quantVal{-} & \quantVal{13.48}
\\
BMBC~\cite{Park:2020:BME} & \quantVal{1.97} & \quantVal{7.66} & \quantVal{-} & \quantVal{13.83} & \quantVal{54.24} & \quantVal{-}
\\
CAIN~\cite{Choi:2020:CAA} & \quantFirst{0.02} & \quantVal{0.06} & \quantVal{0.21} & \quantVal{\hphantom{0}}\quantVal{0.14} & \quantVal{\hphantom{0}}\quantVal{0.39} & \quantVal{\hphantom{0}}\quantVal{1.48}
\\
CtxSyn~\cite{Niklaus:2018:CAS} & \quantVal{0.06} & \quantVal{0.19} & \quantVal{0.76} & \quantVal{\hphantom{0}}\quantVal{0.23} & \quantVal{\hphantom{0}}\quantVal{0.86} & \quantVal{\hphantom{0}}\quantVal{3.49}
\\
DAIN~\cite{Bao:2019:DAV} & \quantVal{0.21} & \quantVal{0.81} & \quantVal{-} & \quantVal{\hphantom{0}}\quantVal{0.68} & \quantVal{\hphantom{0}}\quantVal{2.87} & \quantVal{-}
\\
EDSC~\cite{Cheng:2022:MVF} & \quantFirst{0.02} & \quantVal{0.09} & \quantVal{0.33} & \quantVal{\hphantom{0}}\quantVal{0.17} & \quantVal{\hphantom{0}}\quantVal{0.61} & \quantVal{\hphantom{0}}\quantVal{2.35}
\\
EMA~\cite{Zhang:2023:EMA} & \quantVal{0.07} & \quantVal{0.28} & \quantVal{-} & \quantVal{\hphantom{0}}\quantVal{0.52} & \quantVal{\hphantom{0}}\quantVal{1.98} & \quantVal{-}
\\
FLDR~\cite{Nottebaum:2022:EFE} & \quantVal{0.05} & \quantVal{0.14} & \quantVal{0.59} & \quantVal{\hphantom{0}}\quantVal{0.34} & \quantVal{\hphantom{0}}\quantVal{0.97} & \quantVal{\hphantom{0}}\quantVal{4.00}
\\
M2M~\cite{Hu:2022:MMS} & \quantFirst{0.02} & \quantVal{0.05} & \quantVal{0.17} & \quantVal{\hphantom{0}}\quantVal{0.17} & \quantVal{\hphantom{0}}\quantVal{0.34} & \quantVal{\hphantom{0}}\quantVal{1.18}
\\
RRIN~\cite{Li:2020:VFI} & \quantVal{0.04} & \quantVal{0.13} & \quantVal{0.57} & \quantVal{\hphantom{0}}\quantVal{0.25} & \quantVal{\hphantom{0}}\quantVal{0.91} & \quantVal{\hphantom{0}}\quantVal{3.99}
\\
SepConv~\cite{Niklaus:2017:VFIAS} & \quantVal{0.04} & \quantVal{0.13} & \quantVal{0.46} & \quantVal{\hphantom{0}}\quantVal{0.28} & \quantVal{\hphantom{0}}\quantVal{0.96} & \quantVal{\hphantom{0}}\quantVal{3.17}
\\
SepConv++~\cite{Niklaus:2021:RAC} & \quantFirst{0.02} & \quantVal{0.15} & \quantVal{0.58} & \quantVal{\hphantom{0}}\quantVal{0.16} & \quantVal{\hphantom{0}}\quantVal{1.03} & \quantVal{\hphantom{0}}\quantVal{4.03}
\\
SoftSplat~\cite{Niklaus:2020:SSV} & \quantVal{0.06} & \quantVal{0.21} & \quantVal{0.81} & \quantVal{\hphantom{0}}\quantVal{0.30} & \quantVal{\hphantom{0}}\quantVal{1.08} & \quantVal{\hphantom{0}}\quantVal{4.23}
\\
SplatSyn~\cite{Niklaus:2023:SSV} & \quantVal{0.03} & \quantFirst{0.04} & \quantFirst{0.09} & \quantVal{\hphantom{0}}\quantFirst{0.04} & \quantVal{\hphantom{0}}\quantFirst{0.06} & \quantVal{\hphantom{0}}\quantFirst{0.20}
\\
UPR-Net-L~\cite{Jin:2023:UPR} & \quantVal{0.06} & \quantVal{0.16} & \quantVal{0.61} & \quantVal{\hphantom{0}}\quantVal{0.41} & \quantVal{\hphantom{0}}\quantVal{1.17} & \quantVal{\hphantom{0}}\quantVal{4.30}
\\
VFIformer~\cite{Lu:2022:VFI} & \quantVal{1.26} & \quantVal{4.85} & \quantVal{-} & \quantVal{\hphantom{0}}\quantVal{8.88} & \quantVal{33.77} & \quantVal{-}
\\
VOS~\cite{Yoo:2023:VOS} & \quantFirst{0.02} & \quantVal{0.06} & \quantVal{0.23} & \quantVal{\hphantom{0}}\quantVal{0.11} & \quantVal{\hphantom{0}}\quantVal{0.42} & \quantVal{\hphantom{0}}\quantVal{1.61}
\\
XVFI~\cite{Sim:2021:XVF} & \quantVal{0.05} & \quantVal{0.12} & \quantVal{0.61} & \quantVal{\hphantom{0}}\quantVal{0.34} & \quantVal{\hphantom{0}}\quantVal{0.90} & \quantVal{\hphantom{0}}\quantVal{4.37}
        \\ \bottomrule
    \end{tabularx}\vspace{-0.2cm}
    \captionof{table}{Overview of the runtime during inference in seconds. The missing numbers are due to out-of-memory issues or minimum resolution requirements.}\vspace{-0.0cm}
    \label{tbl:suppl-timetable}
\end{minipage}\qquad\begin{minipage}[b]{.477\textwidth}
    \setlength{\tabcolsep}{0.0cm}
    \renewcommand{\arraystretch}{1.2}
    \newcommand{\quantTit}[1]{\multicolumn{2}{c}{\scriptsize #1}}
    \newcommand{\quantSec}[1]{\scriptsize #1}
    \newcommand{\quantInd}[1]{\scriptsize #1}
    \newcommand{\quantVal}[1]{\scalebox{0.83}[1.0]{$ #1 $}}
    \newcommand{\quantFirst}[1]{\usolid{\scalebox{0.83}[1.0]{$ #1 $}}}
    \newcommand{\quantSecond}[1]{\udotted{\scalebox{0.83}[1.0]{$ #1 $}}}
    \footnotesize
    \begin{tabularx}{\columnwidth}{@{\hspace{0.1cm}} X P{1.25cm} @{\hspace{-0.4cm}} P{1.12cm} P{1.25cm} @{\hspace{-0.4cm}} P{1.12cm} P{1.25cm} @{\hspace{-0.4cm}} P{1.12cm}}
        \toprule
            & \quantTit{no ensemble} & \quantTit{8 $\times$ ensemble} & \quantTit{16 $\times$ ensemble}
        \\ \cmidrule(l{2pt}r{2pt}){2-3} \cmidrule(l{2pt}r{2pt}){4-5} \cmidrule(l{2pt}r{2pt}){6-7}
            & \quantSec{rank} & \quantSec{$\text{PSNR}^{\ast}$} & \quantSec{rank} & \quantSec{$\text{PSNR}^{\ast}$} & \quantSec{rank} & \quantSec{$\text{PSNR}^{\ast}$}
        \\ \midrule
ABME~\cite{Park:2021:ABM} & \quantVal{\hphantom{0}2\text{ / } 20} & \quantVal{28.63} & \quantVal{\hphantom{0}2\text{ / } 20} & \quantVal{29.38} & \quantVal{\hphantom{0}5\text{ / } 20} & \quantVal{29.46}
\\
AdaCoF~\cite{Lee:2020:ACF} & \quantVal{20\text{ / } 20} & \quantVal{20.44} & \quantVal{20\text{ / } 20} & \quantVal{21.86} & \quantVal{19\text{ / } 20} & \quantVal{22.21}
\\
AMT-G~\cite{Li:2023:AMT} & \quantVal{\hphantom{0}7\text{ / } 20} & \quantVal{28.18} & \quantVal{\hphantom{0}6\text{ / } 20} & \quantVal{29.04} & \quantVal{\hphantom{0}3\text{ / } 20} & \quantVal{29.50}
\\
BiFormer~\cite{Park:2023:LBM} & \quantVal{-} & \quantVal{-} & \quantVal{-} & \quantVal{-} & \quantVal{-} & \quantVal{-}
\\
BMBC~\cite{Park:2020:BME} & \quantVal{14\text{ / } 20} & \quantVal{22.18} & \quantVal{14\text{ / } 20} & \quantVal{23.20} & \quantVal{14\text{ / } 20} & \quantVal{23.52}
\\
CAIN~\cite{Choi:2020:CAA} & \quantVal{18\text{ / } 20} & \quantVal{21.83} & \quantVal{17\text{ / } 20} & \quantVal{22.56} & \quantVal{17\text{ / } 20} & \quantVal{22.85}
\\
CtxSyn~\cite{Niklaus:2018:CAS} & \quantVal{\hphantom{0}1\text{ / } 20} & \quantFirst{29.36} & \quantVal{\hphantom{0}1\text{ / } 20} & \quantFirst{30.29} & \quantVal{\hphantom{0}1\text{ / } 20} & \quantFirst{30.44}
\\
DAIN~\cite{Bao:2019:DAV} & \quantVal{\hphantom{0}8\text{ / } 20} & \quantVal{27.83} & \quantVal{\hphantom{0}8\text{ / } 20} & \quantVal{28.77} & \quantVal{\hphantom{0}7\text{ / } 20} & \quantVal{29.09}
\\
EDSC~\cite{Cheng:2022:MVF} & \quantVal{13\text{ / } 20} & \quantVal{22.90} & \quantVal{13\text{ / } 20} & \quantVal{23.47} & \quantVal{13\text{ / } 20} & \quantVal{23.73}
\\
EMA~\cite{Zhang:2023:EMA} & \quantVal{\hphantom{0}9\text{ / } 20} & \quantVal{25.95} & \quantVal{\hphantom{0}9\text{ / } 20} & \quantVal{26.21} & \quantVal{\hphantom{0}9\text{ / } 20} & \quantVal{26.36}
\\
FLDR~\cite{Nottebaum:2022:EFE} & \quantVal{10\text{ / } 20} & \quantVal{24.09} & \quantVal{10\text{ / } 20} & \quantVal{24.81} & \quantVal{11\text{ / } 20} & \quantVal{24.87}
\\
M2M~\cite{Hu:2022:MMS} & \quantVal{\hphantom{0}3\text{ / } 20} & \quantVal{28.61} & \quantVal{\hphantom{0}3\text{ / } 20} & \quantVal{29.32} & \quantVal{\hphantom{0}4\text{ / } 20} & \quantVal{29.48}
\\
RRIN~\cite{Li:2020:VFI} & \quantVal{15\text{ / } 20} & \quantVal{22.17} & \quantVal{15\text{ / } 20} & \quantVal{22.94} & \quantVal{15\text{ / } 20} & \quantVal{23.35}
\\
SepConv~\cite{Niklaus:2017:VFIAS} & \quantVal{16\text{ / } 20} & \quantVal{22.00} & \quantVal{16\text{ / } 20} & \quantVal{22.69} & \quantVal{16\text{ / } 20} & \quantVal{22.89}
\\
SepConv++~\cite{Niklaus:2021:RAC} & \quantVal{17\text{ / } 20} & \quantVal{21.85} & \quantVal{18\text{ / } 20} & \quantVal{22.21} & \quantVal{18\text{ / } 20} & \quantVal{22.24}
\\
SoftSplat~\cite{Niklaus:2020:SSV} & \quantVal{\hphantom{0}4\text{ / } 20} & \quantVal{28.55} & \quantVal{\hphantom{0}3\text{ / } 20} & \quantVal{29.32} & \quantVal{\hphantom{0}2\text{ / } 20} & \quantVal{29.61}
\\
SplatSyn~\cite{Niklaus:2023:SSV} & \quantVal{\hphantom{0}5\text{ / } 20} & \quantVal{28.45} & \quantVal{\hphantom{0}5\text{ / } 20} & \quantVal{29.09} & \quantVal{\hphantom{0}6\text{ / } 20} & \quantVal{29.23}
\\
UPR-Net-L~\cite{Jin:2023:UPR} & \quantVal{\hphantom{0}6\text{ / } 20} & \quantVal{28.31} & \quantVal{\hphantom{0}7\text{ / } 20} & \quantVal{28.84} & \quantVal{\hphantom{0}8\text{ / } 20} & \quantVal{28.95}
\\
VFIformer~\cite{Lu:2022:VFI} & \quantVal{11\text{ / } 20} & \quantVal{23.83} & \quantVal{12\text{ / } 20} & \quantVal{24.53} & \quantVal{10\text{ / } 20} & \quantVal{24.91}
\\
VOS~\cite{Yoo:2023:VOS} & \quantVal{19\text{ / } 20} & \quantVal{20.95} & \quantVal{19\text{ / } 20} & \quantVal{21.89} & \quantVal{20\text{ / } 20} & \quantVal{22.16}
\\
XVFI~\cite{Sim:2021:XVF} & \quantVal{12\text{ / } 20} & \quantVal{23.73} & \quantVal{11\text{ / } 20} & \quantVal{24.65} & \quantVal{12\text{ / } 20} & \quantVal{24.78}
        \\ \bottomrule
    \end{tabularx}\vspace{-0.2cm}
    \captionof{table}{Analysis of the impact of performing ensembling~\cite{Niklaus:2021:RAC} at a 1K resolution. We ask participants to refrain from using it.\\}
    \label{tbl:suppl-ensembling}
\end{minipage}\end{figure*}

\begin{figure*}
    \hspace{-0.18cm}\includegraphics[]{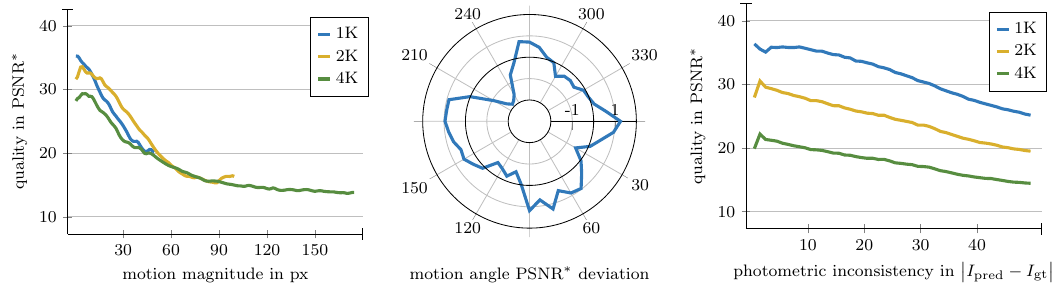}\vspace{-0.2cm}
    \caption{ABME~\cite{Park:2021:ABM} quality in terms of motion magnitude (left), motion angle (center), and photometric inconsistency (right).}\vspace{-0.0cm}
    \label{fig:perpixel-abme}
\end{figure*}

\begin{figure*}
    \hspace{-0.18cm}\includegraphics[]{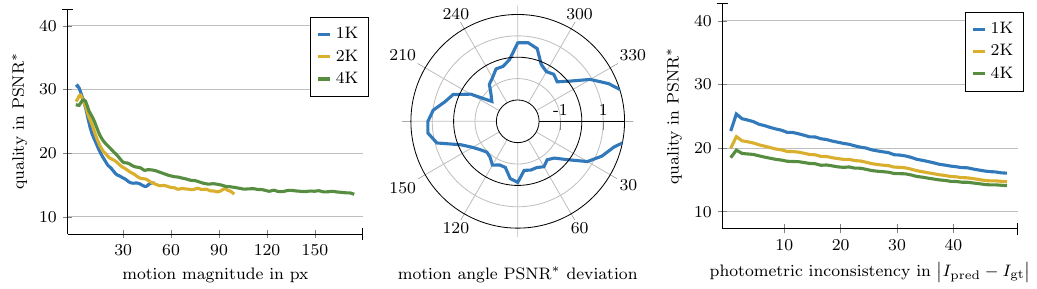}\vspace{-0.2cm}
    \caption{AdaCoF~\cite{Lee:2020:ACF} quality in terms of motion magnitude (left), motion angle (center), and photometric inconsistency (right).}\vspace{-0.0cm}
    \label{fig:perpixel-adacof}
\end{figure*}

\begin{figure*}
    \hspace{-0.18cm}\includegraphics[]{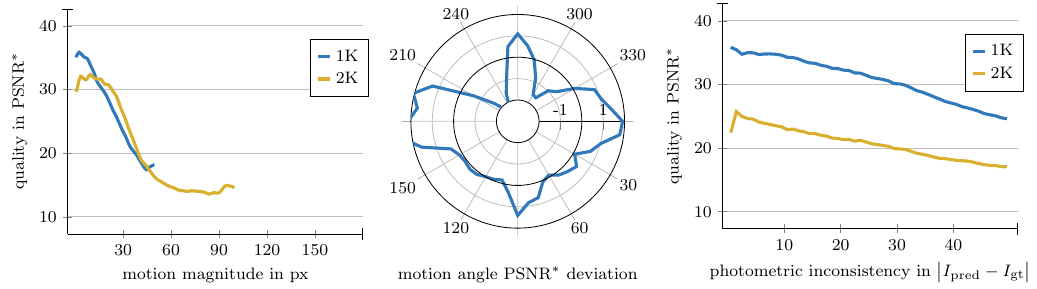}\vspace{-0.2cm}
    \caption{AMT-G~\cite{Li:2023:AMT} quality in terms of motion magnitude (left), motion angle (center), and photometric inconsistency (right).}\vspace{-0.0cm}
    \label{fig:perpixel-amt}
\end{figure*}

\begin{figure*}
    \hspace{-0.18cm}\includegraphics[]{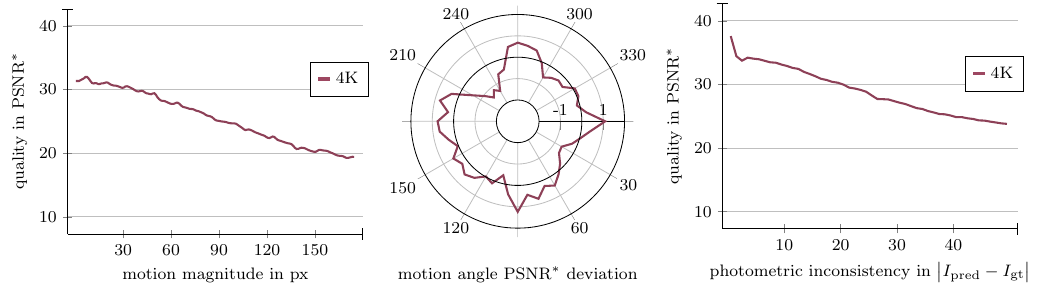}\vspace{-0.2cm}
    \caption{BiFormer~\cite{Park:2023:LBM} quality in terms of motion magnitude (left), motion angle (center), and photometric inconsistency (right). Note that we cannot report all resolutions due to the minimum resolution required by BiFormer.}\vspace{-0.4cm}
    \label{fig:perpixel-biformer}
\end{figure*}

\begin{figure*}
    \hspace{-0.18cm}\includegraphics[]{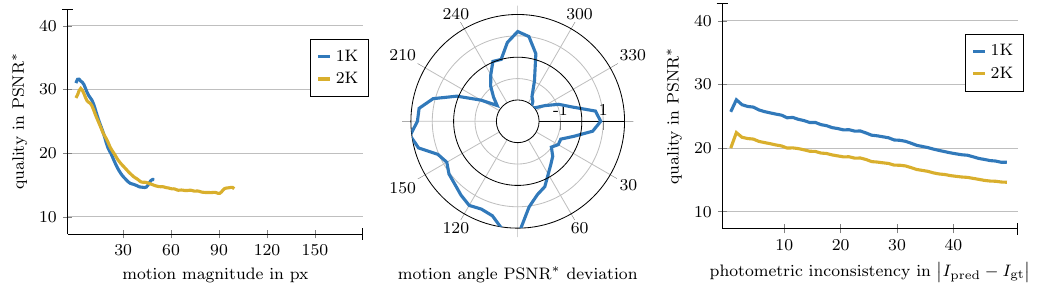}\vspace{-0.2cm}
    \caption{BMBC~\cite{Park:2020:BME} quality in terms of motion magnitude (left), motion angle (center), and photometric inconsistency (right).}\vspace{-0.0cm}
    \label{fig:perpixel-bmbc}
\end{figure*}

\begin{figure*}
    \hspace{-0.18cm}\includegraphics[]{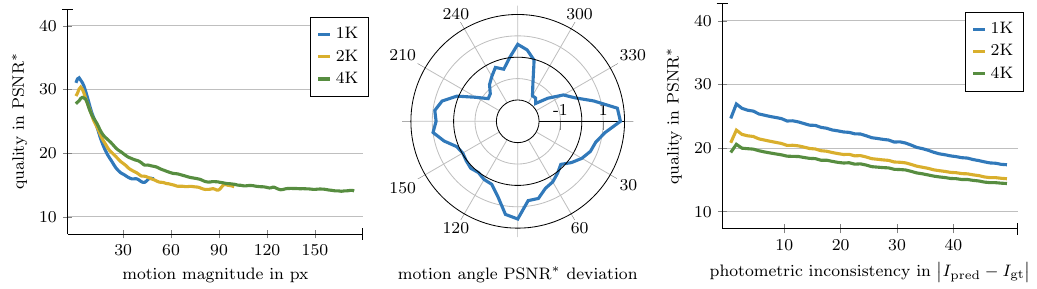}\vspace{-0.2cm}
    \caption{CAIN~\cite{Choi:2020:CAA} quality in terms of motion magnitude (left), motion angle (center), and photometric inconsistency (right).}\vspace{-0.0cm}
    \label{fig:perpixel-cain}
\end{figure*}

\begin{figure*}
    \hspace{-0.18cm}\includegraphics[]{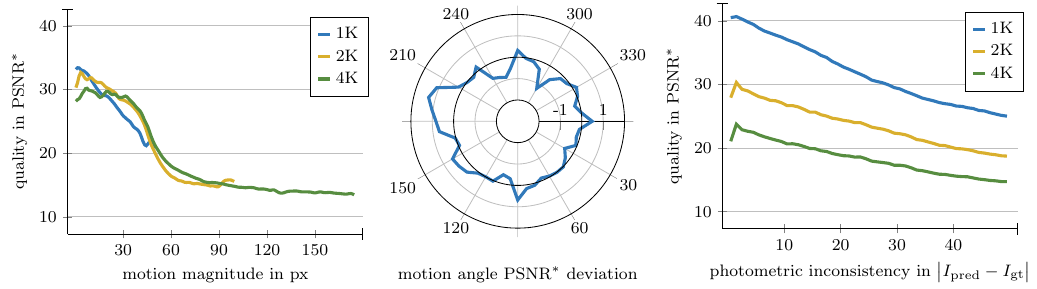}\vspace{-0.2cm}
    \caption{CtxSyn~\cite{Niklaus:2018:CAS} quality in terms of motion magnitude (left), motion angle (center), and photometric inconsistency (right).}\vspace{-0.0cm}
    \label{fig:perpixel-ctxsyn}
\end{figure*}

\begin{figure*}
    \hspace{-0.18cm}\includegraphics[]{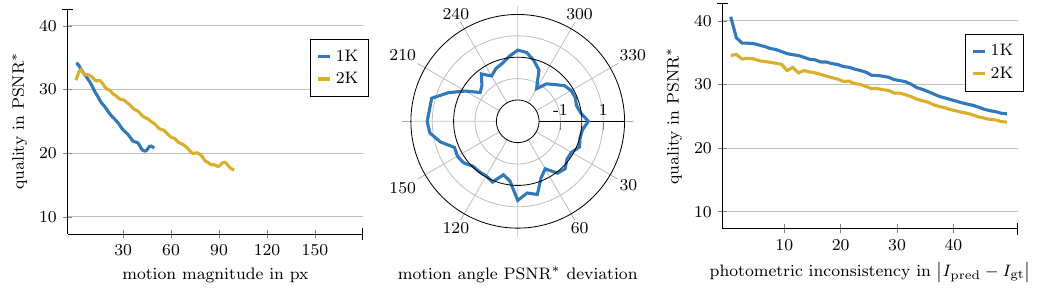}\vspace{-0.2cm}
    \caption{DAIN~\cite{Bao:2019:DAV} quality in terms of motion magnitude (left), motion angle (center), and photometric inconsistency (right).}\vspace{-0.0cm}
    \label{fig:perpixel-dain}
\end{figure*}

\begin{figure*}
    \hspace{-0.18cm}\includegraphics[]{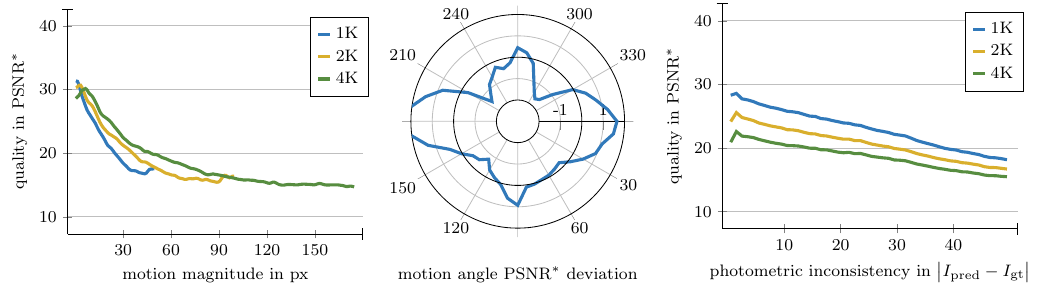}\vspace{-0.2cm}
    \caption{EDSC~\cite{Cheng:2022:MVF} quality in terms of motion magnitude (left), motion angle (center), and photometric inconsistency (right).}\vspace{-0.0cm}
    \label{fig:perpixel-edsc}
\end{figure*}

\begin{figure*}
    \hspace{-0.18cm}\includegraphics[]{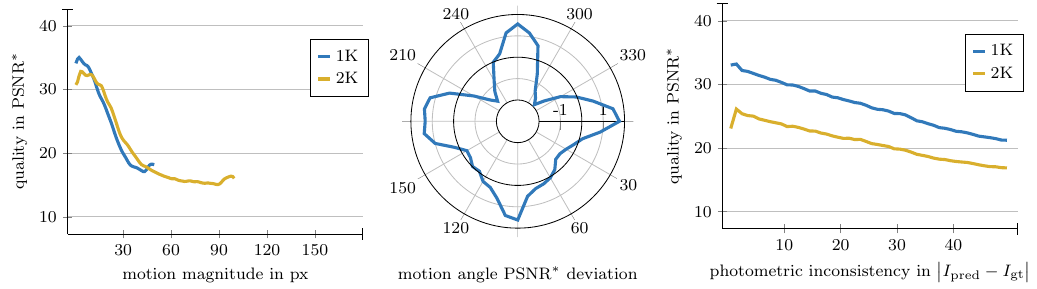}\vspace{-0.2cm}
    \caption{EMA~\cite{Zhang:2023:EMA} quality in terms of motion magnitude (left), motion angle (center), and photometric inconsistency (right).}\vspace{-0.0cm}
    \label{fig:perpixel-ema}
\end{figure*}

\begin{figure*}
    \hspace{-0.18cm}\includegraphics[]{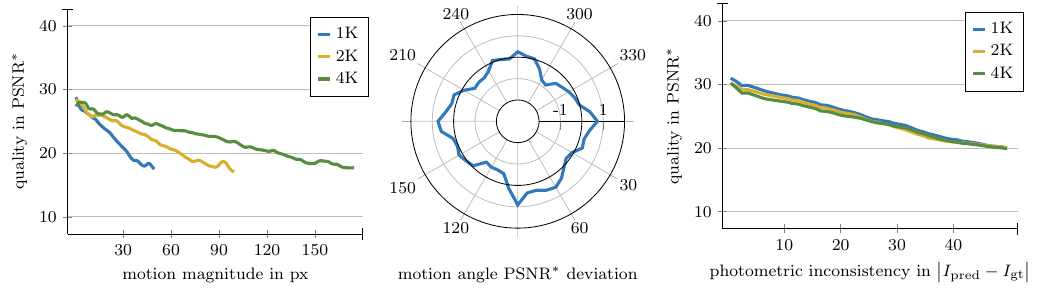}\vspace{-0.2cm}
    \caption{FLDR~\cite{Nottebaum:2022:EFE} quality in terms of motion magnitude (left), motion angle (center), and photometric inconsistency (right).}\vspace{-0.0cm}
    \label{fig:perpixel-fldr}
\end{figure*}

\begin{figure*}
    \hspace{-0.18cm}\includegraphics[]{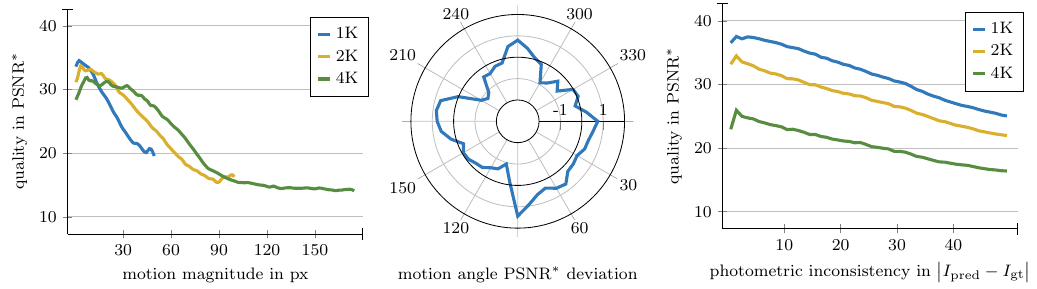}\vspace{-0.2cm}
    \caption{M2M~\cite{Hu:2022:MMS} quality in terms of motion magnitude (left), motion angle (center), and photometric inconsistency (right).}\vspace{-0.0cm}
    \label{fig:perpixel-m2m}
\end{figure*}

\begin{figure*}
    \hspace{-0.18cm}\includegraphics[]{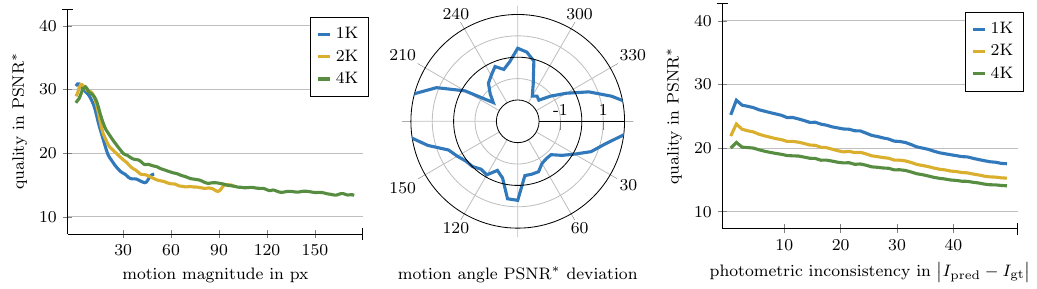}\vspace{-0.2cm}
    \caption{RRIN~\cite{Li:2020:VFI} quality in terms of motion magnitude (left), motion angle (center), and photometric inconsistency (right).}\vspace{-0.0cm}
    \label{fig:perpixel-rrin}
\end{figure*}

\begin{figure*}
    \hspace{-0.18cm}\includegraphics[]{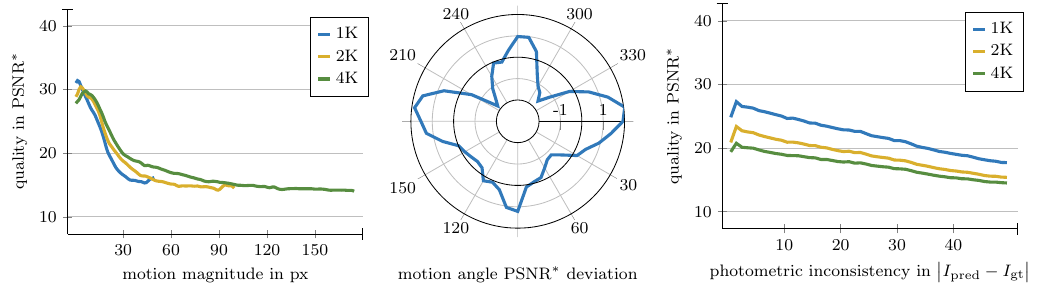}\vspace{-0.2cm}
    \caption{SepConv~\cite{Niklaus:2017:VFIAS} quality in terms of motion magnitude (left), motion angle (center), and photometric inconsistency (right).}\vspace{-0.0cm}
    \label{fig:perpixel-sepconv}
\end{figure*}

\begin{figure*}
    \hspace{-0.18cm}\includegraphics[]{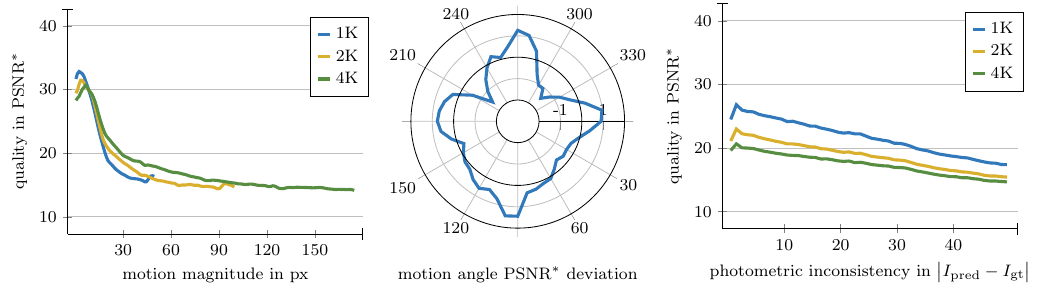}\vspace{-0.2cm}
    \caption{SepConv++~\cite{Niklaus:2021:RAC} quality in terms of motion magnitude (left), motion angle (center), and photometric inconsistency (right).}\vspace{-0.0cm}
    \label{fig:perpixel-sepconvpp}
\end{figure*}

\begin{figure*}
    \hspace{-0.18cm}\includegraphics[]{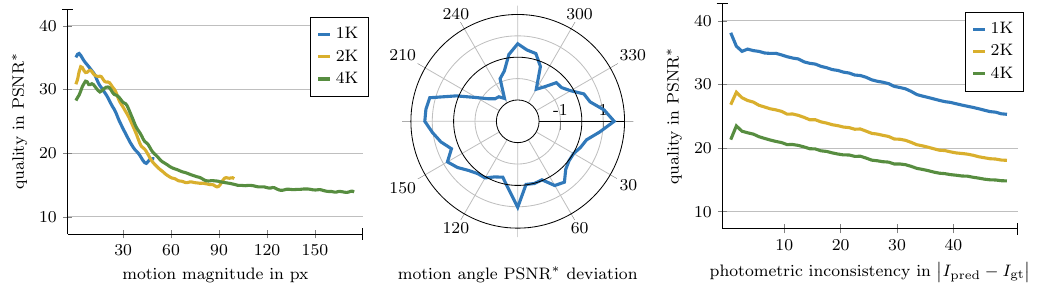}\vspace{-0.2cm}
    \caption{SoftSplat~\cite{Niklaus:2020:SSV} quality in terms of motion magnitude (left), motion angle (center), and photometric inconsistency (right).}\vspace{-0.0cm}
    \label{fig:perpixel-softsplat}
\end{figure*}

\begin{figure*}
    \hspace{-0.18cm}\includegraphics[]{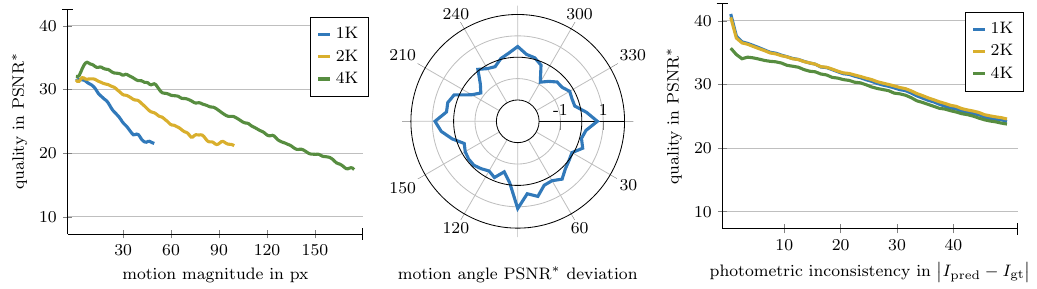}\vspace{-0.2cm}
    \caption{SplatSyn~\cite{Niklaus:2023:SSV} quality in terms of motion magnitude (left), motion angle (center), and photometric inconsistency (right).}\vspace{-0.0cm}
    \label{fig:perpixel-splatsyn}
\end{figure*}

\begin{figure*}
    \hspace{-0.18cm}\includegraphics[]{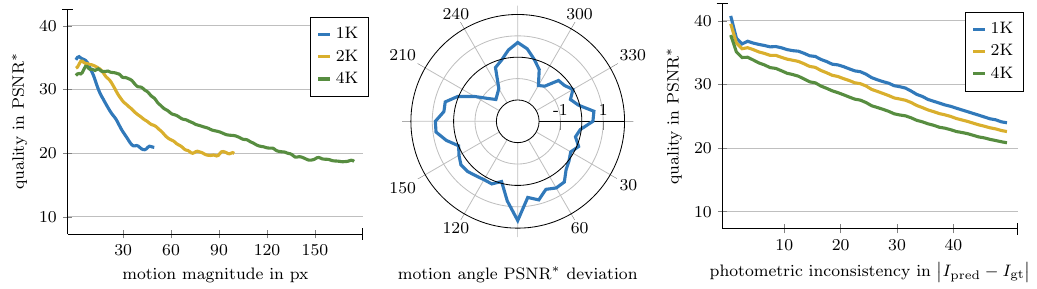}\vspace{-0.2cm}
    \caption{UPR-Net-L~\cite{Jin:2023:UPR} quality in terms of motion magnitude (left), motion angle (center), and photometric inconsistency (right).}\vspace{-0.0cm}
    \label{fig:perpixel-uprnet}
\end{figure*}

\begin{figure*}
    \hspace{-0.18cm}\includegraphics[]{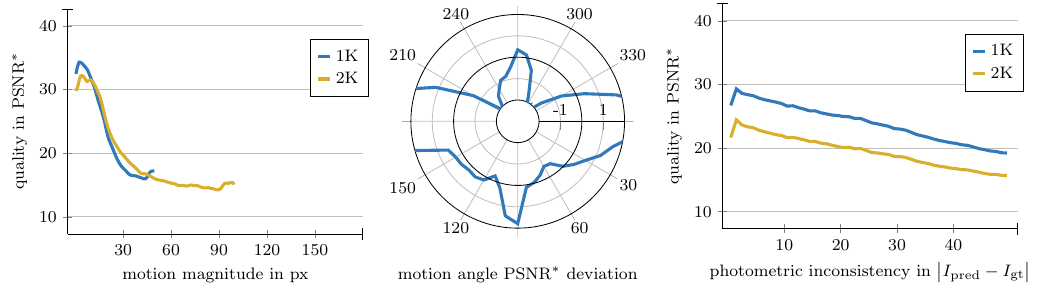}\vspace{-0.2cm}
    \caption{VFIformer~\cite{Lu:2022:VFI} quality in terms of motion magnitude (left), motion angle (center), and photometric inconsistency (right).}\vspace{-0.0cm}
    \label{fig:perpixel-vfiformer}
\end{figure*}

\begin{figure*}
    \hspace{-0.18cm}\includegraphics[]{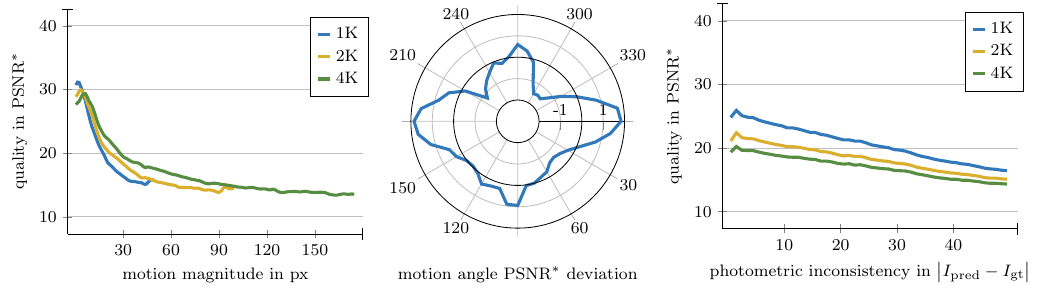}\vspace{-0.2cm}
    \caption{VOS~\cite{Yoo:2023:VOS} quality in terms of motion magnitude (left), motion angle (center), and photometric inconsistency (right).}\vspace{-0.0cm}
    \label{fig:perpixel-vos}
\end{figure*}

\begin{figure*}
    \hspace{-0.18cm}\includegraphics[]{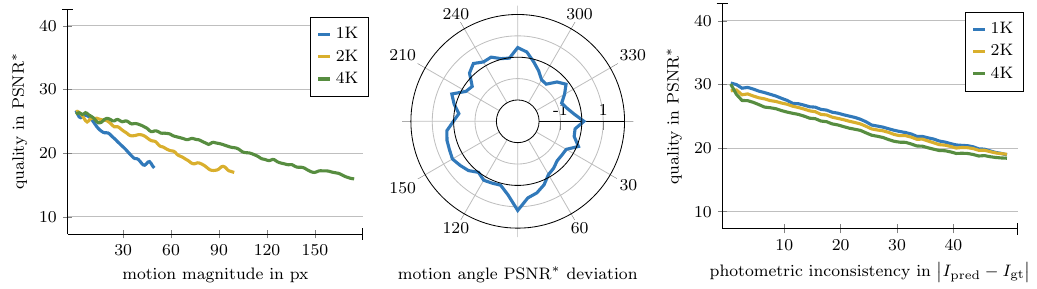}\vspace{-0.2cm}
    \caption{XVFI~\cite{Sim:2021:XVF} quality in terms of motion magnitude (left), motion angle (center), and photometric inconsistency (right).}\vspace{-0.0cm}
    \label{fig:perpixel-xvfi}
\end{figure*}

\begin{figure*}
    \centering
\begin{lstlisting}[basicstyle=\scriptsize]
PyTorch version: 2.2.1+cu121
Is debug build: False
CUDA used to build PyTorch: 12.1

OS: Ubuntu 20.04.6 LTS (x86_64)
GCC version: (Ubuntu 9.4.0-1ubuntu1~20.04.2) 9.4.0
Clang version: 10.0.0-4ubuntu1 
CMake version: version 3.16.3
Libc version: glibc-2.31

Python version: 3.11.5 (main, Sep 11 2023, 13:54:46) [GCC 11.2.0] (64-bit runtime)
Python platform: Linux-5.15.0-1049-aws-x86_64-with-glibc2.31
Is CUDA available: True
CUDA runtime version: 12.1.105
CUDA_MODULE_LOADING set to: LAZY
GPU models and configuration: 
GPU 0: NVIDIA A100-SXM4-40GB
GPU 1: NVIDIA A100-SXM4-40GB
GPU 2: NVIDIA A100-SXM4-40GB
GPU 3: NVIDIA A100-SXM4-40GB
GPU 4: NVIDIA A100-SXM4-40GB
GPU 5: NVIDIA A100-SXM4-40GB
GPU 6: NVIDIA A100-SXM4-40GB
GPU 7: NVIDIA A100-SXM4-40GB

Nvidia driver version: 535.129.03
cuDNN version: Could not collect
HIP runtime version: N/A
MIOpen runtime version: N/A
Is XNNPACK available: True

CPU:
Architecture:                       x86_64
CPU op-mode(s):                     32-bit, 64-bit
CPU(s):                             96
On-line CPU(s) list:                0-95
Thread(s) per core:                 2
Core(s) per socket:                 24
Socket(s):                          2
NUMA node(s):                       2
Model name:                         Intel(R) Xeon(R) Platinum 8275CL CPU @ 3.00GHz
Stepping:                           7
CPU MHz:                            2999.998
Hypervisor vendor:                  KVM
Virtualization type:                full
L1d cache:                          1.5 MiB
L1i cache:                          1.5 MiB
L2 cache:                           48 MiB
L3 cache:                           71.5 MiB
NUMA node0 CPU(s):                  0-23,48-71
NUMA node1 CPU(s):                  24-47,72-95

Versions of relevant libraries:
[conda] numpy                     1.26.2                   pypi_0    pypi
[conda] torch                     2.2.1+cu121              pypi_0    pypi
[conda] torchmetrics              1.2.0                    pypi_0    pypi
[conda] torchvision               0.17.1+cu121             pypi_0    pypi
\end{lstlisting}\vspace{-0.4cm}
    \caption{Environment for measuring the inference runtime (from PyTorch's \texttt{get\_pretty\_env\_info}), only GPU 0 was used.}
    \label{fig:environment}
\end{figure*}